\documentclass[runningheads, envcountsame, a4paper]{llncs}

\usepackage[T1]{fontenc}

\usepackage[hyphens]{url}

\usepackage{hyperref}
\hypersetup{hidelinks,
  colorlinks=true,
  allcolors=black,
  pdfstartview=Fit,
  breaklinks=true}
\usepackage[all]{hypcap}

\usepackage[group-four-digits,per-mode=fraction]{siunitx}
\usepackage[pdftex,dvipsnames]{xcolor}  % Coloured text etc.

\usepackage[misc]{ifsym}

\usepackage{cite}
\usepackage{enumitem}

\usepackage[pdftex]{graphicx}
\usepackage{epstopdf}
\usepackage[caption = false]{subfig}

\usepackage{amsmath}
\usepackage{amssymb}

\usepackage{csquotes}

\usepackage{microtype}

\usepackage{tikz}
\usetikzlibrary{matrix,positioning,arrows.meta,arrows,fit,backgrounds,decorations.pathreplacing}

\tikzset{
	mymat/.style={
		matrix of math nodes,
		text height=2.5ex,
		text depth=0.75ex,
		text width=6.00ex,
		align=center,
		column sep=-\pgflinewidth,
		nodes={minimum height=5.0ex}
	},
	mymats/.style={
		mymat,
		nodes={draw,fill=#1}
	}, 
	mymat2/.style={
		matrix of math nodes,
		text height=1.0ex,
		text depth=0.0ex,
		minimum width=5ex,
		align=center,
		column sep=-\pgflinewidth
	},	
}

\newcommand{\R}{\mathbb{R}}

\newcommand{\Ours}{Consequence-aware Sequential Counterfactuals}
\newcommand{\ours}{\textsc{cscf}}
\newcommand{\oursShort}{\ours}
\newcommand{\oursBoth}{\textsc{(c)scf}}

\newcommand{\mainEA}{\oursShort}
\newcommand{\competitor}{\textsc{synth}}

\newcommand{\alternativeEA}{\textsc{scf}}

\newcommand{\adultDataset}{\emph{Adult Census}}
\newcommand{\germanDataset}{\emph{German Credit}}

\newcommand{\actions}{\mathcal{A}}
\newcommand{\action}{a}
\newcommand{\actionspart}{\mathbb{A}}
\newcommand{\Values}{\mathcal{V}}
\newcommand{\Value}{v}
\newcommand{\valuespart}{\mathbb{V}}

\newcommand{\state}{\mathbf{x}}

\newcommand{\initialinstance}{\state_{0}}
\newcommand{\originalclass}{\texttt{reject}}
\newcommand{\targetclass}{\texttt{accept}}

\newcommand{\featurespace}{\mathcal{X}}
\newcommand{\feature}{\dddot{\featurespace}}
\newcommand{\featureindex}{\mathcal{I}}

\newcommand{\predictionspace}{\mathcal{Y}}

\newcommand{\transition}{\tau}

\newcommand{\constraint}{\mathbb{C}}

\newcommand{\sequence}{\mathcal{S}}

\newcommand{\objective}{o}
\newcommand{\seqCosts}{\objective_{1}}
\newcommand{\seqDistance}{\objective_{2}}

\newcommand{\genotype}{G}
\newcommand{\phenotype}{P}
\newcommand{\decoder}{D}

\newcommand{\blackbox}{f}

\newcommand{\dependencyGraph}{\mathcal{G}}
\newcommand{\consequentialCosts}{g}
\newcommand{\baseCosts}{b}
\newcommand{\costs}{c}
\newcommand{\totalCosts}{C}

\newcommand{\finalsolution}{\state_{T}}

\begin{document}

%opening
\title{Consequence-aware Sequential Counterfactual Generation}
\toctitle{Consequence-aware Sequential Counterfactual Generation}

\author{Philip~Naumann\inst{1,2} (\Letter) \and
Eirini~Ntoutsi\inst{1,2}}
\tocauthor{Philip~Naumann and Eirini~Ntoutsi}
\authorrunning{P.~Naumann and E.~Ntoutsi}
\institute{Freie Universit{\"a}t Berlin, Germany \and
{L3S} Research Center, Leibniz Universit{\"a}t Hannover, Germany \\%\and
% Leibniz Universit{\"a}t Hannover, Hannover, Germany \\
\email{$\{$philip.naumann, eirini.ntoutsi$\}$@fu-berlin.de}}

\maketitle
\setcounter{footnote}{0}

\begin{abstract}
Counterfactuals have become a popular technique nowadays for interacting with black-box machine learning models and understanding how to change a particular instance to obtain a desired outcome from the model. However, most existing approaches assume instant materialization of these changes, ignoring that they may require effort and a specific order of application. Recently, methods have been proposed that also consider the order in which actions are applied, leading to the so-called sequential counterfactual generation problem.

In this work, we propose a model-agnostic method for sequential counterfactual generation. We formulate the task as a multi-objective optimization problem and present a genetic algorithm approach to find optimal sequences of actions leading to the counterfactuals. Our cost model considers not only the direct effect of an action, but also its consequences.
Experimental results show that compared to state-of-the-art, our approach generates less costly solutions, is more efficient and provides the user with a diverse set of solutions to choose from.

\end{abstract}

\section{Introduction}
\label{sec:introduction}
Due to the increasing use of machine learning algorithms in sensitive areas such as law, finance or labor, there is an equally increased need for transparency and so-called \emph{recourse} options~\cite{wachterCounterfactualExplanationsOpening2017,karimiSurveyAlgorithmicRecourse2020}.
It is no longer sufficient to simply deliver a decision, but moreover to be able to explain it and, ideally, offer assistance if one feels unfairly treated by the algorithm.
Hiding the decision-making algorithm behind a (virtual) wall like an API makes these issues especially intransparent and problematic in case of \emph{black-box} models, as they are not able to communicate with the end user beyond the provided decision (e.g. \originalclass{} or \targetclass{}).

For this reason, algorithms emerged that aim to explain a (black-box) decision or even provide essential recourse information in order to change an undesired outcome in one's favor.
The latter of these methods is of particular interest, since it has the capability to improve a bad decision for someone into a good one by giving explicit directions on what to change with respect to the provided information. These methods are commonly referred to as \emph{counterfactual explanations}~\cite{wachterCounterfactualExplanationsOpening2017}. The goal here is to change (tweak) characteristics (features) of a provided input so that the black-box decision turns out in favor afterwards (e.g. increase your level of education to get a higher salary). The result of applying these changes on the input is commonly referred to as a \emph{counterfactual}~\cite{wachterCounterfactualExplanationsOpening2017}.

Usually, those changes are atomic operations, meaning each feature is tweaked independently~\cite{wachterCounterfactualExplanationsOpening2017,mothilalExplainingMachineLearning2020,dandlMultiObjectiveCounterfactualExplanations2020a,laugelComparisonBasedInverseClassification2018}. Thus, feature interrelationships, e.g. of causal nature, are not considered.
Recent approaches propose to define changes in (multiple) features through so-called \emph{actions}, which can be thought of as instructions on how to apply the modifications and their consequences (e.g. increasing the level of education has an impact on age because it takes time to obtain a degree).
Actions further help to describe what features are actionable, mutable or immutable~\cite{karimiSurveyAlgorithmicRecourse2020}. However, these approaches, like the traditional counterfactual methods, still assume that all these changes happen instantly and do not consider implications and consequences of the \emph{order} of their application.

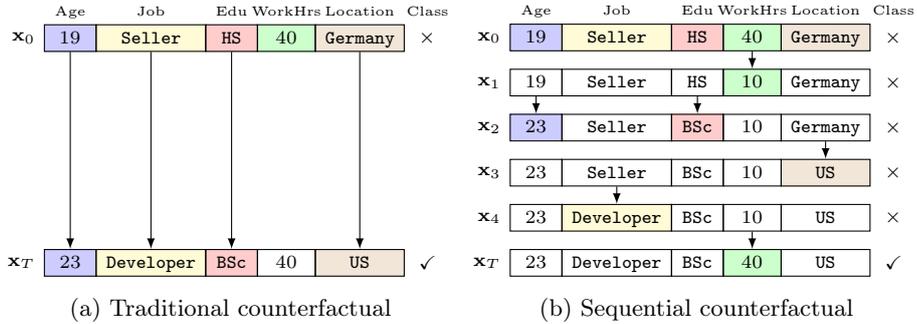
\begin{figure}[!ht]
	\centering
	\subfloat[Traditional counterfactual]{\begin{tikzpicture}[>=latex, nodes={font=\scriptsize}]
	\matrix[mymat2,anchor=west,nodes={draw},
	column 1/.style={nodes={text width=2ex}},
	column 2/.style={nodes={text width=9ex}},
	column 3/.style={nodes={text width=3ex}},
	column 4/.style={nodes={text width=4ex}},
	column 5/.style={nodes={text width=7ex}},]
	at (0,0) 
	(x0)
	{   
		19 & \texttt{Seller} & \texttt{HS} & 40  & \texttt{Germany} \\
	};

% \phantom{
% 	\matrix[mymat2,below=0.0ex of x0,nodes={draw},
% 		column 1/.style={nodes={text width=2ex}},
% 	column 2/.style={nodes={text width=9ex}},
% 	column 3/.style={nodes={text width=3ex}},
% 	column 4/.style={nodes={text width=4ex}},
% 	column 5/.style={nodes={text width=7ex}},]
% 	(x1)
% 	{   
% 		19 & \texttt{Seller} & \texttt{HS} & 10  & \texttt{US} \\ 
% 	};}
% \phantom{
% 	\matrix[mymat2,below=0.0ex of x1,nodes={draw},
% 		column 1/.style={nodes={text width=2ex}},
% 	column 2/.style={nodes={text width=9ex}},
% 	column 3/.style={nodes={text width=3ex}},
% 	column 4/.style={nodes={text width=4ex}},
% 	column 5/.style={nodes={text width=7ex}},]
% 	(x2)
% 	{   
% 		23 & \texttt{Seller} & \texttt{BSc} & 10  & \texttt{US} \\ 
% 	};}
% \phantom{
% 	\matrix[mymat2,below=0.0ex of x2,nodes={draw},
% 		column 1/.style={nodes={text width=2ex}},
% 	column 2/.style={nodes={text width=9ex}},
% 	column 3/.style={nodes={text width=3ex}},
% 	column 4/.style={nodes={text width=4ex}},
% 	column 5/.style={nodes={text width=7ex}},]
% 	(x3)
% 	{   
% 		23 & \texttt{Seller} & \texttt{BSc} & 10  & \texttt{US} \\ 
% 	};}
% \phantom{
% 	\matrix[mymat2,below=0.0ex of x3,nodes={draw},
% 		column 1/.style={nodes={text width=2ex}},
% 	column 2/.style={nodes={text width=9ex}},
% 	column 3/.style={nodes={text width=3ex}},
% 	column 4/.style={nodes={text width=4ex}},
% 	column 5/.style={nodes={text width=7ex}},]
% 	(x4)
% 	{   
% 		23 & \texttt{Developer} & \texttt{BSc} & 10  & \texttt{US} \\ 
% 	};}

	\matrix[mymat2,below=17.5ex of x0,nodes={draw},
	column 1/.style={nodes={text width=2ex}},
	column 2/.style={nodes={text width=9ex}},
	column 3/.style={nodes={text width=3ex}},
	column 4/.style={nodes={text width=4ex}},
	column 5/.style={nodes={text width=7ex}},]
	(xT)
	{
		23 & \texttt{Developer} & \texttt{BSc} & 40 & \texttt{US} \\
	};
	
	\node[above=0pt of x0-1-1.north, text depth=0pt]
	(a1) {\tiny Age};
	\node[above=0pt of x0-1-2.north, text depth=0pt]
	(a2) {\tiny Job};
	\node[above=0pt of x0-1-3.north, text depth=0pt]
	(a3) {\tiny Edu};
	\node[above=0pt of x0-1-4.north, text depth=0pt]
	(a4) {\tiny WorkHrs};	
	\node[above=0pt of x0-1-5.north, text depth=0pt]
	(a5) {\tiny Location};	

    \node[left=0pt of x0-1-1.west]
    {$\initialinstance$};
    \node[left=0pt of xT-1-1.west]
    {$\finalsolution$};
    
    \node[right=-2pt of a5.east, text depth=0pt]
    {\tiny Class};
    \node[right=2pt of x0-1-5.east]
    {$\times$};
    \node[right=2pt of xT-1-5.east]
    {\checkmark};
	
	\draw[->]
		(x0-1-1.south) -- (xT-1-1.north);
	\draw[->]
		(x0-1-2.south) -- (xT-1-2.north);
	\draw[->]
		(x0-1-3.south) -- (xT-1-3.north);
	\draw[->]
	    (x0-1-5.south) -- (xT-1-5.north);

\begin{scope}[on background layer]
  \node [fill=blue!20,inner xsep=0pt, inner ysep=0pt] [fit=(x0-1-1)] {};
  \node [fill=yellow!20,inner xsep=0pt, inner ysep=0pt] [fit=(x0-1-2)] {};
  \node [fill=red!20,inner xsep=0pt, inner ysep=0pt] [fit=(x0-1-3)] {};
  \node [fill=green!20,inner xsep=0pt, inner ysep=0pt] [fit=(x0-1-4)] {};
  \node [fill=brown!20,inner xsep=0pt, inner ysep=0pt] [fit=(x0-1-5)] {};
  
  \node [fill=blue!20,inner xsep=0pt, inner ysep=0pt] [fit=(xT-1-1)] {};
  \node [fill=yellow!20,inner xsep=0pt, inner ysep=0pt] [fit=(xT-1-2)] {};
  \node [fill=red!20,inner xsep=0pt, inner ysep=0pt] [fit=(xT-1-3)] {};
  \node [fill=brown!20,inner xsep=0pt, inner ysep=0pt] [fit=(xT-1-5)] {};
 \end{scope}
	
\end{tikzpicture}}
	\subfloat[Sequential counterfactual]{\begin{tikzpicture}[>=latex, nodes={font=\scriptsize}]
	\matrix[mymat2,anchor=west,nodes={draw},
		column 1/.style={nodes={text width=2ex}},
	column 2/.style={nodes={text width=9ex}},
	column 3/.style={nodes={text width=3ex}},
	column 4/.style={nodes={text width=4ex}},
	column 5/.style={nodes={text width=7ex}},]
	at (0,0) 
	(x0)
	{   
		19 & \texttt{Seller} & \texttt{HS} & 40 & \texttt{Germany} \\ 
	};
	
	\matrix[mymat2,below=0.0ex of x0,nodes={draw},
		column 1/.style={nodes={text width=2ex}},
	column 2/.style={nodes={text width=9ex}},
	column 3/.style={nodes={text width=3ex}},
	column 4/.style={nodes={text width=4ex}},
	column 5/.style={nodes={text width=7ex}},]
	(x1)
	{   
		19 & \texttt{Seller} & \texttt{HS} & 10  & \texttt{Germany} \\ 
	};
	
	\matrix[mymat2,below=0.0ex of x1,nodes={draw},
		column 1/.style={nodes={text width=2ex}},
	column 2/.style={nodes={text width=9ex}},
	column 3/.style={nodes={text width=3ex}},
	column 4/.style={nodes={text width=4ex}},
	column 5/.style={nodes={text width=7ex}},]
	(x2)
	{   
		23 & \texttt{Seller} & \texttt{BSc} & 10  & \texttt{Germany} \\ 
	};
	
	\matrix[mymat2,below=0.0ex of x2,nodes={draw},
		column 1/.style={nodes={text width=2ex}},
	column 2/.style={nodes={text width=9ex}},
	column 3/.style={nodes={text width=3ex}},
	column 4/.style={nodes={text width=4ex}},
	column 5/.style={nodes={text width=7ex}},]
	(x3)
	{   
		23 & \texttt{Seller} & \texttt{BSc} & 10  & \texttt{US} \\ 
	};
	
	\matrix[mymat2,below=0.0ex of x3,nodes={draw},
		column 1/.style={nodes={text width=2ex}},
	column 2/.style={nodes={text width=9ex}},
	column 3/.style={nodes={text width=3ex}},
	column 4/.style={nodes={text width=4ex}},
	column 5/.style={nodes={text width=7ex}},]
	(x4)
	{   
		23 & \texttt{Developer} & \texttt{BSc} & 10  & \texttt{US} \\ 
	};

	\matrix[mymat2,below=0.0ex of x4,nodes={draw},
		column 1/.style={nodes={text width=2ex}},
	column 2/.style={nodes={text width=9ex}},
	column 3/.style={nodes={text width=3ex}},
	column 4/.style={nodes={text width=4ex}},
	column 5/.style={nodes={text width=7ex}},]
	(xT)
	{
		23 & \texttt{Developer} & \texttt{BSc} & 40  & \texttt{US} \\
	};
	
	\node[above=0pt of x0-1-1.north, text depth=0pt]
	(a1) {\tiny Age};
	\node[above=0pt of x0-1-2.north, text depth=0pt]
	(a2) {\tiny Job};
	\node[above=0pt of x0-1-3.north, text depth=0pt]
	(a3) {\tiny Edu};
	\node[above=0pt of x0-1-4.north, text depth=0pt]
	(a4) {\tiny WorkHrs};	
	\node[above=0pt of x0-1-5.north, text depth=0pt]
	(a5) {\tiny Location};		

    \node[left=0pt of x0-1-1.west]
    {$\initialinstance$};
    \node[left=0pt of x1-1-1.west]
    {$\state_1$};
    \node[left=0pt of x2-1-1.west]
    {$\state_2$};
    \node[left=0pt of x3-1-1.west]
    {$\state_3$};
    \node[left=0pt of x4-1-1.west]
    {$\state_4$};
    \node[left=0pt of xT-1-1.west]
    {$\finalsolution$};
	
    \node[right=-2pt of a5.east, text depth=0pt]
    {\tiny Class};
    \node[right=2pt of x0-1-5.east]
    {$\times$};
    \node[right=2pt of x1-1-5.east]
    {$\times$};
    \node[right=2pt of x2-1-5.east]
    {$\times$};
    \node[right=2pt of x3-1-5.east]
    {$\times$};
    \node[right=2pt of x4-1-5.east]
    {$\times$};
    \node[right=2pt of xT-1-5.east]
    {\checkmark};

	\draw[->]
		(x0-1-4.south) -- (x1-1-4.north);
		
	\draw[->]
		(x1-1-1.south) -- (x2-1-1.north);
	\draw[->]
		(x1-1-3.south) -- (x2-1-3.north);
		
	\draw[->]
		(x2-1-5.south) -- (x3-1-5.north);
		
	\draw[->]
		(x3-1-2.south) -- (x4-1-2.north);

	\draw[->]
		(x4-1-4.south) -- (xT-1-4.north);

\begin{scope}[on background layer]
  \node [fill=blue!20,inner xsep=0pt, inner ysep=0pt] [fit=(x0-1-1)] {};
  \node [fill=yellow!20,inner xsep=0pt, inner ysep=0pt] [fit=(x0-1-2)] {};
  \node [fill=red!20,inner xsep=0pt, inner ysep=0pt] [fit=(x0-1-3)] {};
  \node [fill=green!20,inner xsep=0pt, inner ysep=0pt] [fit=(x0-1-4)] {};
  \node [fill=brown!20,inner xsep=0pt, inner ysep=0pt] [fit=(x0-1-5)] {};

  \node [fill=green!20,inner xsep=0pt, inner ysep=0pt] [fit=(x1-1-4)] {};

  \node [fill=blue!20,inner xsep=0pt, inner ysep=0pt] [fit=(x2-1-1)] {};
  \node [fill=red!20,inner xsep=0pt, inner ysep=0pt] [fit=(x2-1-3)] {};
  
  \node [fill=brown!20,inner xsep=0pt, inner ysep=0pt] [fit=(x3-1-5)] {};
  
  \node [fill=yellow!20,inner xsep=0pt, inner ysep=0pt] [fit=(x4-1-2)] {};
  
  \node [fill=green!20,inner xsep=0pt, inner ysep=0pt] [fit=(xT-1-4)] {};
\end{scope}

\end{tikzpicture}} 
	\caption{The difference between traditional counterfactual generation (a) and the sequential approach (b). Although the generated counterfactual $\finalsolution$ is the same, the process and implied knowledge/information is different.}
	\label{fig:sequential_vs_traditional}
\end{figure}

For this reason, recent works regard the application of feature altering changes as a \emph{sequential} process~\cite{ramakrishnanSynthesizingActionSequences2020,shavitExtractingIncentivesBlackBox2019}.
The problem is then shifted from simply computing the counterfactual, to finding an ordered \emph{sequence} of actions that accomplishes it. In Fig.~\ref{fig:sequential_vs_traditional} we visualized this difference for a person $\initialinstance$ wishing to attain a higher salary (Class: $\times \Rightarrow \checkmark$) for which getting a higher education level, switching the job and changing the location is necessary. In this case, e.g., it was assumed that decreasing one's working hours is \emph{beneficial} in order to obtain a higher degree, which is a relationship that traditional approaches do not model. Each state $\state_t$ in Fig.~\ref{fig:sequential_vs_traditional}b describes the result of applying an action on the previous state $\state_{t-1}$ and $\finalsolution$ denotes the counterfactual.
As we can see, the actions have different effects in the feature space and may alter more than one feature at a time. Moreover, the order makes a difference as increasing the level of education before changing the job is usually more beneficial, as well as decreasing the working hours in order to attain the degree (representing a focus on the education). Later on, this change is reset to its original value through another action since an increased value is now more \emph{plausible} again with respect to the job change.
The whole process of Fig.~\ref{fig:sequential_vs_traditional}b can be seen as a \emph{sequential counterfactual} as each state is an important part of it, whereas in the traditional setting in Fig.~\ref{fig:sequential_vs_traditional}a only the final result is important. Thus, a sequential counterfactual allows us to look beyond just flipping the class label and provides further information about the underlying process. It is not the absolute main goal anymore to only switch the class label, but to take consequential effects into account in order to improve the overall benefit of the actions. For the end user, this also means getting concrete information on the actions and their order.
Our work is in the direction of sequential counterfactual generation and inspired by~\cite{ramakrishnanSynthesizingActionSequences2020}, which we will further elaborate in Sect.~\ref{sec:related_work}.

The rest of this paper is organized as follows: we will first give an overview on related work in Sect.~\ref{sec:related_work}. Then, we introduce the \emph{consequence-aware counterfactual generation problem} in Sect.~\ref{sec:problem}, for which our proposed method follows in Sect.~\ref{sec:method}. Lastly, we evaluate it in Sect.~\ref{sec:experiments} to a state-of-the-art method and give final conclusions in Sect.~\ref{sec:conclusion}.
We want to note that the metaphorical examples used throughout do not always correspond to reality and are merely illustrative.

\section{Related Work}
\label{sec:related_work}
Counterfactual explanations were first introduced in~\cite{wachterCounterfactualExplanationsOpening2017} as a mean towards algorithmic transparency. Motivated by the \enquote{closest possible world}~\cite{wachterCounterfactualExplanationsOpening2017} in which a favorable outcome is true (e.g. \targetclass{}), most methods optimize on the \emph{distance} between the original input $\initialinstance$ and the counterfactual $\finalsolution$ to keep the changes minimal. Since this notion alone was found to be insufficient, other popular objectives include the \emph{plausibility} (often measured as the counterfactual being within the class distribution~\cite{joshiRealisticIndividualRecourse2019} or being close to instances from the dataset~\cite{poyiadziFACEFeasibleActionable2020,dandlMultiObjectiveCounterfactualExplanations2020a,vanlooverenInterpretableCounterfactualExplanations2019}) and \emph{sparsity}~\cite{dandlMultiObjectiveCounterfactualExplanations2020a,mothilalExplainingMachineLearning2020,vanlooverenInterpretableCounterfactualExplanations2019} (measuring how many features had to change) of solutions.
Desirable criteria regarding the algorithms are, e.g., being model-agnostic~\cite{dandlMultiObjectiveCounterfactualExplanations2020a,karimiModelAgnosticCounterfactualExplanations2020,lashGeneralizedInverseClassification2017,laugelComparisonBasedInverseClassification2018} (e.g. by using genetic algorithms) or providing a diverse set of solutions~\cite{mothilalExplainingMachineLearning2020,dandlMultiObjectiveCounterfactualExplanations2020a,downsCRUDSCounterfactualRecourse2020,mahajanPreservingCausalConstraints2020} (e.g. by using multi-objective optimization).

More recently, works~\cite{ustunActionableRecourseLinear2019,karimiAlgorithmicRecourseImperfect2020a,ramakrishnanSynthesizingActionSequences2020} began to replace the distance function with a \emph{cost} in order to express aspects such as the \emph{effort} of a change. 
In alignment to this, the term \emph{recourse}~\cite{joshiRealisticIndividualRecourse2019,ustunActionableRecourseLinear2019,karimiAlgorithmicRecourseImperfect2020a,karimiSurveyAlgorithmicRecourse2020,poyiadziFACEFeasibleActionable2020,downsCRUDSCounterfactualRecourse2020} has attracted attention to describe the counterfactual generation.
It can be defined as \enquote{the ability of a person to change the decision of a model by altering actionable input variables}~\cite{ustunActionableRecourseLinear2019} and thus emphasizes on the \emph{actionability} of the features to provide comprehensible recommendations that can be acted upon~\cite{karimiSurveyAlgorithmicRecourse2020}.
In addition, more attention has been paid on the causal nature of feature interrelationships, e.g. by including causal models in the generation process to assess mutual effects~\cite{downsCRUDSCounterfactualRecourse2020,karimiAlgorithmicRecourseImperfect2020a,mahajanPreservingCausalConstraints2020}.

Lastly, motivated by the fact that in reality most changes do not happen instantly, but are rather part of a \emph{process}, there have been works that extend the formulation of actions and their consequences by incorporating them in a sequential setting~\cite{ramakrishnanSynthesizingActionSequences2020,shavitExtractingIncentivesBlackBox2019}. In contrast to simply finding the counterfactual that switches the class label, the focus is on finding a subset of actions that, applied in a specific order, accomplishes the counterfactual while accounting for potential consequences of prior actions on subsequent ones (cf. Fig.~\ref{fig:sequential_vs_traditional}).

In this work, we also focus on sequential counterfactual generation. The advantages of our method in comparison to~\cite{ramakrishnanSynthesizingActionSequences2020} can be summarized as follows: our method is model-agnostic and not bound to differentiable (cost) functions.
It finds diverse sequences instead of a single solution, thus giving more freedom of choice to the end user. Moreover, it is efficient in pruning and exploring the search space due to using already found knowledge (exploitation) and the ability to optimize all sub-problems (cf. Sect.~\ref{sec:moo_problem}) at once, while~\cite{ramakrishnanSynthesizingActionSequences2020} breaks these down into separate problems. This efficiency allows us to find sequences of any length, whereas~\cite{ramakrishnanSynthesizingActionSequences2020} requires multiple runs and more time for it (cf. Sect.~\ref{sec:experiments}).
Another difference is our action-cost model (cf. Sect.~\ref{sec:costs}). We regard consequences not only in the feature space (e.g. age increases as a consequence of obtaining a higher degree), but also explicitly model their effects in the objective or cost space (e.g. changing the job becomes \emph{easier} with a higher degree).
This way we extend the cost formulation of~\cite{ramakrishnanSynthesizingActionSequences2020}, which only proposes to model (feature) relationships through (boolean) pre- and post-requirement constraints (e.g. you \emph{must} at least be 18 to get a driver's license). These constraints are also possible in our model. 
Finally, we note that the work of~\cite{shavitExtractingIncentivesBlackBox2019} also discusses consequential effects. However, since no cost function is optimized, but instead the target class likelihood of the counterfactual, we do not compare with~\cite{shavitExtractingIncentivesBlackBox2019} in this work.

\section{Problem Statement}
\label{sec:problem}
We assume a static \emph{black-box classifier} $\blackbox: \featurespace \to \predictionspace$ where  $\featurespace = \featurespace_1 \times \cdots \times \featurespace_d$ is the feature space and $\predictionspace$ is the class space.
The notation $\feature_h$ is used to refer to the feature itself (e.g. $\feature_{\texttt{Edu}}$ denotes \texttt{Education}, whereas $\featurespace_{\texttt{Edu}} = \{\dots, \texttt{HS}, \texttt{BSc}, \dots\}$ is the domain).
For simplicity and without loss of generality, we assume $\blackbox$ is a binary classifier with $\predictionspace=\{\originalclass,\targetclass\}$. 
Let $\initialinstance \in \featurespace$ be an instance of the problem, e.g. a person seeking to receive an annual salary of more than \$50k, and the current decision of $\blackbox$ based on $\initialinstance$ is $\blackbox(\initialinstance) = \originalclass$. The goal is to find a counterfactual example $\finalsolution$ for $\initialinstance$ such that $\blackbox(\finalsolution) = \targetclass$. 
In other words, we want to change the original instance so that it will receive a positive decision from the black-box. The sort of changes we refer to are in the feature description of the instance, like increasing \texttt{Age} by 5 years or decreasing \texttt{Work Hours} to 20 per week. 
Our problem formulation builds upon~\cite{ramakrishnanSynthesizingActionSequences2020}. 
We introduce actions in Sect.~\ref{sec:actions} along with their associated cost to implement the suggested changes in Sect.~\ref{sec:costs}.
Finally, we formulate the generation of sequential counterfactuals as a multi-objective optimization problem in Sect.~\ref{sec:moo_problem}.

\subsection{Actions, Sequences of Actions and States}
\label{sec:actions}
Let $\actions = \{\action_1,\dots,\action_n\}$ be a problem-specific, manually-defined set of \emph{actions}.
Each action is a function $\action_i: \featurespace \times \Values \to \featurespace$ that modifies a subset of features $\featureindex_{\action_{i}} = \{ \feature_h, \feature_k, \dots \} \subseteq \feature$ in a given input instance $\state_{t-1} \in \featurespace$ in order to realize a new instance (which we refer to as a \emph{state}) $\action_i(\state_{t-1}, \Value_i) = \state_t \in \featurespace$.
An action \emph{directly} affects one feature $\feature_h \in \featureindex_{\action_{i}}$ (e.g. \texttt{Education}) based on a tweaking value $\Value_i \in \Values_h \subseteq \featurespace_h$ and may have \emph{indirect} effects on other features $\feature_{k \neq h} \in \featureindex_{\action_{i}}$ as a consequence (e.g. \texttt{Age}).
Here, $\Values \subseteq \featurespace$ describes the \emph{feasible value space} that restricts the tweaking values based on the given $\state$ (e.g. \texttt{Age} may only increase).

For example, $\state_2$ in Fig.~\ref{fig:sequential_vs_traditional}b is the result of applying $\action_{\texttt{Edu}}(\state_1, \texttt{BSc}) = \state_2$ which changes the \texttt{Education} ($\texttt{HS} \Rightarrow \texttt{BSc}$) and affects the \texttt{Age} ($19 \Rightarrow 23$) as a consequence. In this case, $\Values_{\texttt{Edu}}$ and $\Values_{\texttt{Age}}$ restrict the tweaking values so that they can only increase.
It is possible to use a causal model as in~\cite{karimiAlgorithmicRecourseImperfect2020a} for evaluating how the indirectly affected features have to be changed.
An action-value pair $(\action_i, \Value_i)$ thus represents a specification how $\featureindex_{\action_i}$ of $\state$ is affected with respect to the tweaking value $\Value_i$ of feature $\feature_h$. 

Additionally, each action $\action_i$ may be subject to boolean constraints $\constraint_i: \featurespace \times \Values \to \mathbb{B}$ such as pre- and post-requirements as proposed in~\cite{ramakrishnanSynthesizingActionSequences2020} (cf. Sect.~\ref{sec:related_work}), which can also be used to validate the feasibility of indirectly affected features. Each action-value pair is considered \emph{valid}, if it satisfies the associated constraints and $\Values$.
An ordered sequence $\sequence$ of valid action-value pairs $(\action_i^t, \Value_i)$ leading to the counterfactual $\finalsolution$ is called a \emph{sequential counterfactual}. Here, $t \in \{1,\dots,T\}$ is the order of applying the actions and $T \in \{1,\dots,|\actions|\}$ is the number of used actions in $\sequence$, i.e. the sequence length.

\subsection{Consequence-aware Cost Model}
\label{sec:costs}
Our goal is to assess the direct effort (which can, e.g., be abstract, as based on personal preferences, or concrete, like money or time etc.) of an action while considering possible consequences. We define the cost of an action $\action_i$ as a function of two components: $\costs_i(\cdot) = \baseCosts_i(\cdot) \cdot \consequentialCosts_i(\cdot)$. Here, $\baseCosts_i$ represents the \emph{direct effort}, whereas $\consequentialCosts_i$ acts as a discount of it in order to express (beneficial) \emph{consequences} of prior actions.
Please note that each action has its \emph{own} cost function.
Summing up all action costs of a sequence $\sequence$ yields the \emph{sequence cost}: $\totalCosts_\sequence = \sum_{\action_i \in \sequence} \costs_i(\cdot)$.
In the following we will explain the components in more detail.

\textbf{Action Effort} $\baseCosts_i$: \,
First, we introduce $\baseCosts_i: \featurespace \times \featurespace \to \R_+$, which is assumed to be an action-specific measure of the direct effort caused by an action $\action_i$. Therefore, this is specified as a function between two consecutive states $\state_{t-1}, \state_t \in \featurespace$, representing the direct effect of applying that action on $\state_{t-1}$. This function can be thought of as a typical cost function as in e.g.~\cite{ramakrishnanSynthesizingActionSequences2020}.
As an example, $\baseCosts_i$ could be specified linear and time based, whereby an effort caused by an action \texttt{addEdu} would be represented by the years required (e.g. four to progress from \texttt{HS} to \texttt{BSc}). Alternatively, monetary costs could be used (i.e. tuition costs), or a combination of both.
Besides, there are no particular conditions on this function, so it can be defined arbitrarily (e.g. return a constant value).

\begin{figure}[!ht]
	\centering
	\subfloat[Feature relationship graph $\dependencyGraph$]{\begin{tikzpicture}[> = latex]
	\begin{scope}[every node/.style={circle,thick,draw}]%,label distance=-3px}]
		\node (a1)[label={[shift={(-0.2,-0.45)}]\tiny $\baseCosts_1 = 10$}] at (0,0) {$\feature_{1}$};
		\node (a2)[label={[shift={(0.12,-0.35)}]\tiny $\baseCosts_2 = 5$}] at (3,0) {$\feature_{2}$};
		\node (a3)[label={[shift={(0.0,-0.35)}]\tiny $\baseCosts_3 = 15$}] at (1.5,2) {$\feature_{3}$};
% 		\node (a1)[label={[shift={(-0.2,-0.45)}]\tiny $\baseCosts_1 = 10$}] at (0,0) {$\feature_{\texttt{Job}}$};
% 		\node (a2)[label={[shift={(0.12,-0.35)}]\tiny $\baseCosts_2 = 5$}] at (3,0) {$\feature_{\texttt{Edu}}$};
% 		\node (a3)[label={[shift={(0.0,-0.35)}]\tiny $\baseCosts_3 = 15$}] at (1.5,2) {$\feature_{\texttt{Loc}}$};
	\end{scope}

	\begin{scope}[
		every node/.style={fill=white,circle},
		every edge/.style={draw=black!60,dashed,very thick}]
		
% 		\path[->] (a1) edge[bend left=30] node {\small $\transition_{12}$} (a2);
		\path[->] (a3) edge node[] { $\transition_{32}$} (a2);
		\path[->] (a2) edge node[] { $\transition_{21}$} (a1);
		\path[->] (a3) edge node[] { $\transition_{31}$} (a1);
	\end{scope}

% 	\begin{sope}
% 	   % \node (1) at (1.0,2.7) {\scriptsize $\baseCosts_3 = $};
% 	   % \node (2) at (-0.1,1.02) {\scriptsize $\consequentialCosts_{1}(e_{31}) = $};
% 	\end{sope}
\end{tikzpicture}} 
	\subfloat[Different sequences $\sequence_1$ (red) and $\sequence_2$ (blue)]{\begin{tikzpicture}[> = latex]

	\begin{scope}[every node/.style={rectangle,thick,draw}]%,label distance=-20px}]
		\node (x0) at (0,1.0) {$\initialinstance$};
		
%		\node (A) at (2,0) {$\action_3$};
		\node (C)[label=above:{\tiny $\costs_3 = \overbrace{15}^{\baseCosts_3} \cdot \overbrace{1.0}^{\consequentialCosts_3}$}] at (1.3,2) {$\action_3^1(\initialinstance)$};
		\node (CC)[label=above:{\tiny $\costs_1 = \overbrace{10}^{\baseCosts_1} \cdot \overbrace{0.75}^{\consequentialCosts_1}$}] at (3.3,2) {$\action_1^2(\state_1)$};
% \node (CC)[label=above:{\tiny $\costs_1 = \overbrace{10}^{\baseCosts_1} \cdot \overbrace{\frac{.5+1}{2}=0.75}^{\consequentialCosts_1}$}] at (3.3,2) {$\action_1^2(\state_1)$};
        \node (xT1)[label=above:{\tiny $\costs_2 = \overbrace{5}^{\baseCosts_2} \cdot \overbrace{1.0}^{\consequentialCosts_2}$}] at (5.3,2) {$\action_2^3(\state_2)$};

		\node (B)[label=below:{\tiny $\costs_2 = \underbrace{5}_{\baseCosts_2} \cdot \underbrace{0.5}_{\consequentialCosts_2}$}] at (1.3,0) {$\action_2^1(\initialinstance)$};
		\node (CCC)[label=below:{\tiny $\costs_3 = \underbrace{15}_{\baseCosts_3} \cdot \underbrace{1.0}_{\consequentialCosts_3}$}] at (3.3,0) {$\action_3^2(\state'_1)$};
        \node (xT2)[label=below:{\tiny $\costs_1 = \underbrace{10}_{\baseCosts_1} \cdot \underbrace{0.5}_{\consequentialCosts_1}$}] at (5.3,0) {$\action_1^3(\state'_2)$};

		\node (xT) at (6.6,1.0) {$\finalsolution$};

	\end{scope}

	\begin{scope}[
% 		every node/.style={fill=white,circle},
		every edge/.style={very thick}]
		
		\draw [->,draw=red!60, very thick,double] (x0) |- node[below, align=center, pos=0.7](dummy1){$\initialinstance$} (C);
		\draw [->,draw=red!60, very thick,double] (C) -- node[below, align=center]{$\state_1$} (CC);
		\draw [->,draw=red!60, very thick,double] (CC) -- node[below, align=center]{$\state_2$} (xT1);
		\draw [->,draw=red!60, very thick, double] (xT1) -| node[below, align=center, pos=0.3](dummy2){$\finalsolution$} (xT);
		
		\draw [->,draw=blue!60, very thick,double] (x0) |- node[above, align=center, pos=0.7](dummy3){$\initialinstance$} (B);
% 		\draw (x0) -- (B) (x) edge[->] (n3.south) (x) edge["EXAMPLE"] (y);
	    \draw [->,draw=blue!60, very thick,double] (B) -- node[above, align=center]{$\state'_1$} (CCC);
		\draw [->,draw=blue!60, very thick,double] (CCC) -- node[above, align=center]{$\state'_2$} (xT2);
		\draw [->,draw=blue!60, very thick, double] (xT2) -| node[above, align=center, pos=0.3](dummy4){$\finalsolution$} (xT);
% 		\path [->] (AA) edge[draw=blue!60] node {\small $\totalCosts=27$} (xT);

% 		\path [->] (BB) edge[draw=red!60] node {\small $\totalCosts=34$} (xT);
		
% 		\draw [->] (x0) edge[draw=black!40,dashed] node[above, align=center] {\scriptsize $\langle \action_i^t, \dots \rangle$} (xT);
		
% 		\node (res)[align=center] at (7.5,1) {$\totalCosts_{\sequence_1} = 27.5$\\$\totalCosts_{\sequence_2} = 22.5$};

	\end{scope}

  \begin{scope}[on background layer]
    % \node (bk1) [fill=red!10,rounded corners, draw=black!50, dashed, inner xsep=25pt, inner ysep=2pt] [fit=(C) (CC) (xT1) ] {};

    % \node (bk2) [fill=blue!10,rounded corners, draw=black!50, dashed, inner xsep=25pt, inner ysep=2pt] [fit=(B) (CCC) (xT2) ] {};
    
    % \node (bk2) [back group] [fit=(p4) (p5)] {};
    % \node (bk3) [back group] [fit=(p6) (p7)] {};
    % \node [draw, thick, green!50!black, fill=green!75!black!25, rounded corners, fit=(p1), inner xsep=15pt, inner ysep=10pt] {};
  \end{scope}
  
    \node [left=0.0ex of dummy1] {$\sequence_1$};
    \node [left=0.0ex of dummy3] {$\sequence_2$};
  
\draw [decorate,
	decoration = {brace,amplitude=5pt,mirror,raise=11pt}] (C.west) --  (xT1.east) node[midway,yshift=-2.1em] {\tiny Cost of $\sequence_1$: $\totalCosts_{\sequence_1}=\costs_3+\costs_1+\costs_2=27.5$};
	
\draw [decorate,
	decoration = {brace,amplitude=5pt,raise=11pt}] (B.west) --  (xT2.east) node[midway,yshift=2.1em] {\tiny Cost of $\sequence_2$: $\totalCosts_{\sequence_2}=\costs_2+\costs_3+\costs_1=22.5$};
	
\end{tikzpicture}} 
	
	\caption{For simplicity, the $\transition(\cdot)$ functions in (a) are based on binary conditions:
	$\transition_{32} = 1.0 \ \text{if} \ \featurespace_3 := \texttt{US}, \ \text{else} \ 0.5$.
	$\transition_{31} = 0.5 \ \text{if} \ \featurespace_3 := \texttt{US}, \ \text{else} \ 1.0$.
	$\transition_{21} = 0.5 \ \text{if} \ \featurespace_2 \geq \texttt{BSc}, \ \text{else} \ 1.0$.
	As a reference, the action efforts $\baseCosts_i$ are provided above each feature in (a).
	}
	\label{fig:action_dependencies}
\end{figure}
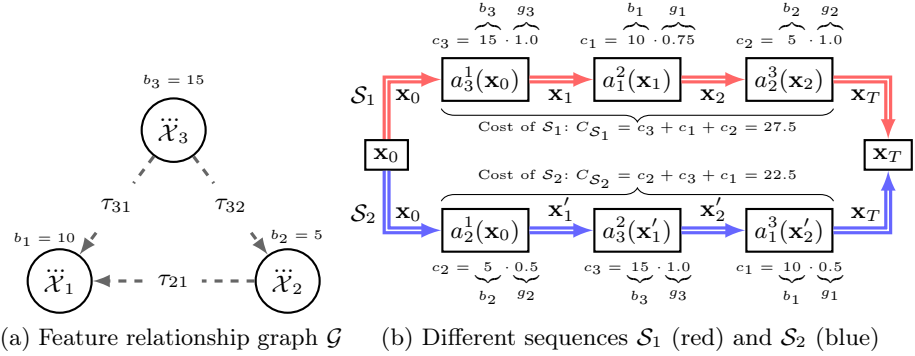

\textbf{Consequential Discount} $\consequentialCosts_i$: \,
To assess a possible (beneficial) consequential effect of previous actions on applying the current one $\action^t_i$, we introduce a so-called \emph{consequential discount} $\consequentialCosts_i: \featurespace \to [0,1]$ that affects the action effort $\baseCosts_i$ based on the current state $\state_{t-1}$ (i.e. before applying $\action^t_i$). Such effects can be, e.g., \enquote{the higher the \texttt{Education}, the easier it is to increase \texttt{Capital}} or \enquote{increasing \texttt{Education} in \texttt{Germany} is cheaper than in the \texttt{US} (due to lower tuition fees)}.
This discount therefore describes a value in $[0,1]$, where 0 would mean that the current state is so beneficial that the effort of the action to be applied is completely cancelled out, and 1 that there is no advantageous effect.
We derive the aforementioned consequential effect on an action from consequential relationships between feature pairs. This is provided as a graph $\dependencyGraph = (X,E)$ where the nodes $X \subseteq \feature$ are a subset of the features and edges $e_{kh} \in E$ between each two nodes $\feature_k, \feature_h \in X$ describe a function $\transition_{kh}: \featurespace \to [0,1]$ that models a consequential effect between one feature $\feature_k$ to another $\feature_h$. For the given features $\feature_1:=\texttt{Job}$, $\feature_2:=\texttt{Education}$ and $\feature_3:=\texttt{Location}$ we have exemplified $\dependencyGraph$ in Fig.~\ref{fig:action_dependencies}a by the following relations:
\begin{enumerate}
    \item The \texttt{Education} cost depends on the \texttt{Location} ($\feature_3 \xrightarrow{\transition_{32}} \feature_2$). E.g., it is \emph{cheaper} to get a degree in \texttt{Germany} than the \texttt{US} because of lower tuition fees.
    \item The easiness of getting a \texttt{Job} depends on the \texttt{Location} ($\feature_3 \xrightarrow{\transition_{31}} \feature_1$). E.g., it is \emph{easier} to get a \texttt{Developer} job in the \texttt{US} than in other locations.
    \item The higher the \texttt{Education}, the \emph{easier} it is to change the \texttt{Job} ($\feature_2 \xrightarrow{\transition_{21}} \feature_1$).
\end{enumerate}

Based on this modeling in $\dependencyGraph$, we can then derive the consequential discount $\consequentialCosts_i$ of an action $\action_i$ (Eq.~\ref{eq:avg_action_discount}) by averaging the consequential effect $\hat{\consequentialCosts}_h$ of \emph{each affected feature} $\feature_h \in \featureindex_{\action_i}$ (Eq.~\ref{eq:avg_feature_discount}).
It is assumed that $\consequentialCosts_i$ evaluates to $1.0$ if no feature in $\featureindex_{\action_i}$ is influenced by another one in $\dependencyGraph$ (e.g. $\feature_3$ in Fig.~\ref{fig:action_dependencies}a).
\begin{align}
  \hat{\consequentialCosts}_h(\state_{t-1}) = \texttt{avg}(\{\transition_{kh}(\state_{t-1}) \ \forall \ \feature_k \in X \mid \exists \ e_{kh} \in E \}) \label{eq:avg_feature_discount}\\
  \consequentialCosts_i(\state_{t-1}) = \texttt{avg}(\{ \hat{\consequentialCosts}_h(\state_{t-1}) \ \forall \ \feature_h \in \featureindex_{\action_i} \mid \exists \ \feature_h \in X \}) \label{eq:avg_action_discount}
\end{align}

In order to understand the benefit of the consequential discount on the sequence order, we exemplify in Fig.~\ref{fig:action_dependencies}b the situation from Fig.~\ref{fig:sequential_vs_traditional}b (for simplicity the working hours altering actions are omitted). The available actions are thus: \enquote{change \texttt{Job} to \texttt{Developer}} ($\action_1$), \enquote{get a \texttt{BSc} degree} ($\action_2$) and \enquote{change \texttt{Location} to \texttt{US}} ($\action_3$) (notice all $\Values_i$ are fixed to a single value). Each $\action_i$ alters their respective feature counterpart $\feature_i$ (e.g. $\featureindex_{\action_1} = \{\feature_1\}$). We can see the cost computations in Fig.~\ref{fig:action_dependencies}b for two differently ordered sequences $\sequence_1 = \langle \action^1_3, \action^2_1, \action^3_2 \rangle$ and $\sequence_2 = \langle \action^1_2, \action^2_3, \action^3_1 \rangle$ that achieve the same final outcome $\finalsolution$ (i.e. only the application order is different).
To compute the consequential discount for action $\action^2_1$ in $\sequence_1$, e.g., we consider the relations $\feature_3 \xrightarrow{\transition_{31}} \feature_1$ and $\feature_2 \xrightarrow{\transition_{21}} \feature_1$ with respect to $\state_1$ to derive the \emph{feature} discount (Eq.~\ref{eq:avg_feature_discount}): $\hat{\consequentialCosts}_{1}(\state_{1}) = \frac{0.5+1.0}{2} = 0.75$. Since no other feature is affected by $\action_1$ according to $\featureindex_{\action_1}$, the \emph{action} discount (Eq.~\ref{eq:avg_action_discount}) evaluates to $\consequentialCosts_1(\state_1) = \hat{\consequentialCosts}_{1}(\state_1)$.
After computing all action costs $\costs_i$, we can derive the sequence costs $\totalCosts_{\sequence_1} = 27.5$ and $\totalCosts_{\sequence_2} = 22.5$, which shows that $\sequence_2$ would be preferred here as it benefits more from the consequential discount effects of $\dependencyGraph$.
Note, that if we leave out the consequential discounts completely, i.e. $\costs_i = \baseCosts_i$, then there would be no notion of order here as each sequence would receive the same costs (assuming the same tweaking values).
Furthermore, our consequence-aware formulation means, that additional actions are only used if their induced effort is lower than the consequential benefit they provide (as this would otherwise make $\totalCosts_\sequence$ worse than if the action was not used).

\subsection{Consequence-aware Sequential Counterfactual Generation}
\label{sec:moo_problem}
Based on the previous definitions, we now introduce the \emph{consequence-aware sequential counterfactual generation} problem. Find the counterfactual $\finalsolution \in \featurespace$ of an initial instance $\initialinstance \in \featurespace$ by taking valid action-value pairs $(\action_i^t, \Value_i)$ of a sequence $\sequence$ according to the constraints $\constraint_i$ and sequence cost $\totalCosts_\sequence$ such that $\blackbox(\finalsolution) = \targetclass$. In order to solve this, we identify three sub-problems:
\begin{enumerate}[label=\textbf{Problem~\arabic*} (Prob.~\arabic*):, ref=Prob.~\arabic*, align=left]
    \item\label{prob:1} Find an \textbf{optimal subset of actions} $\actions^* \subseteq \actions$.
    \item\label{prob:2} Find \textbf{optimal values} $\Values^* \in \Values$ for $\actions^*$.
    \item\label{prob:3} Find the \textbf{optimal order} of $\sequence$ to apply the actions.
\end{enumerate}

For an arbitrary set of actions $\actions$ and feasible value space $\Values$ many sequences can be generated, therefore it is important to assess their quality. For this purpose, we will use the sequence cost $\seqCosts := \totalCosts_\sequence$ as a \emph{subjective} measure, as well as the Gower's distance~\cite{gowerGeneralCoefficientSimilarity1971} $\seqDistance := \texttt{dist}(\initialinstance,\finalsolution)$ to act as an \emph{objective} assessment how much $\finalsolution$ differs from $\initialinstance$. The Gower's distance is able to combine numerical and categorical features and is thus an appropriate measure here~\cite{dandlMultiObjectiveCounterfactualExplanations2020a}. The reason for using both is, that $\seqCosts$ measures the \emph{effort} of the \emph{whole process}, whereas $\seqDistance$ only considers the \emph{difference} to the final counterfactual and is agnostic of the process.

In order to propose diverse solutions, we will formulate the problem as a multi-objective one and add, next to $\seqCosts$ and $\seqDistance$, the individual tweaking frequencies of each of the $1 \leq h \leq d$ features after unrolling the complete sequence, i.e. $\objective_{2+h} = \#(\feature_h \in \featureindex_{\action_i} \ \forall \ \action_i \in \sequence)$. In other words, $\objective_{2+h}$ measures how often a feature $\feature_h$ was affected by all actions of $\sequence$ combined. E.g., the frequencies for $\objective_3, \dots, \objective_7$ would be $\{1,1,1,2,1\}$ in case of Fig.~\ref{fig:sequential_vs_traditional}b.
In a way this can be thought of as the \emph{sparsity} objective mentioned in Sect.~\ref{sec:related_work}, but aggregated individually per feature instead of a sum.
The idea behind the diversity objectives is to keep the number of feature changes minimal and additionally force the optimization to seek alternative changes instead.
This means, solutions mainly compete with those that change the same features with respect to $\seqCosts$ and $\seqDistance$, resulting in a diverse set of optimal options.
Combining all the above yields the following multi-objective minimization problem:
\begin{equation}
    \begin{split}
	\min_{\sequence} & {(\underbrace{\seqCosts}_{\text{Sequence cost}}, \underbrace{\seqDistance}_{\text{Gower's distance}}, \quad \underbrace{\objective_{2+1}, \dots, \objective_{2+h} , \dots, \objective_{2+d}}_{\text{Feature tweaking frequencies}})} \\ \text{s.t.} \ & \blackbox(\finalsolution) = \targetclass \ \ \text{and} \bigwedge_{(\action_i, \Value_i) \in \sequence}{\constraint_i} \\
	\end{split}
	\label{eq:problem}
\end{equation}

% \section{Method}
\section{\Ours{} (CSCF)}
\label{sec:method}
In order to address the combinatorial (\ref{prob:1}, \ref{prob:3}) and continuous (\ref{prob:2}) sub-problems with respect to Eq.~\ref{eq:problem}, we used a \emph{Biased Random-Key Genetic Algorithm} (BRKGA)~\cite{goncalvesBiasedRandomkeyGenetic2011} and adapted it for multi-objective optimization by using non-dominated sorting (NDS)~\cite{srinivasMuiltiobjectiveOptimizationUsing1994}. 
NDS is preferred over a scalarization approach to avoid manual prioritization of the objectives and to address them equally.
Moreover, by using BRKGA we avoid the manual definition of problem-specific operators, which is not trivial here.
This choice allows to solve all sub-problems at once, is model-agnostic, derivative-free and provides multiple solutions.
% Other genetic algorithms are also possible, but defining suitable operators is not trivial and BRKGA already offers all key advantages we need.
% Note that it is also possible to use other genetic algorithms. However, defining suitable operators is not trivial and BRKGA already offers all key advantages we need.

\textbf{BRKGA}: \,
The main idea behind BRKGA is that it optimizes on the genotype of the solutions and evaluates on their phenotype, making the optimization itself problem-independent~\cite{goncalvesBiasedRandomkeyGenetic2011}. A \emph{genotype} is the internal representation of a solution, whereas the \emph{phenotype} is the actual solution we wish to generate. The phenotype must always be deterministically derivable from the genotype through a \emph{decoder} function~\cite{goncalvesBiasedRandomkeyGenetic2011}.
Because of this, each solution genotype in BRKGA is encoded as a vector of real (random) values in the range of $[0,1]$ (the so-called \emph{random-keys})~\cite{goncalvesBiasedRandomkeyGenetic2011}.

In each generation (iteration), the decoded solution phenotypes are evaluated based on their fitness (given by the vector of evaluating each objective of Eq.~\ref{eq:problem} individually) and the population is divided into two subsets by applying NDS: the so-called \emph{elites}, which are the feasible (valid), non-dominated (i.e. best) solutions in the Pareto-front~\cite{eibenIntroductionEvolutionaryComputing2015}, and the remaining ones, called \emph{non-elites}.
Then, genetic mating is performed by selecting two parents, one from the elite sub-population and one from the non-elites. A new solution is created by applying biased crossover~\cite{goncalvesBiasedRandomkeyGenetic2011} on the two parents, which favors selecting the value of the elite solution with a certain biased probability. This step is repeated until sufficient new solutions have been created. Additionally, a number of completely random solutions are generated in order to preserve the exploration of the search space and the diversity in the population. Finally, the different sub-populations (\emph{elites}, \emph{crossovered} and \emph{random} solutions) are merged and evaluated and form the new population for the next generation. This loop continues until some termination criterion is reached. The Pareto-front of the last generation then represents our final solution set. Note that the Pareto-front usually holds more than one solution (i.e. a diverse set of optimal sequences according to the objectives of Eq.~\ref{eq:problem}).

\textbf{Genotype}: \,
We wish to solve the three sub-problems from Sect.~\ref{sec:moo_problem} at once.
Inspired by similar, successful, encodings of problems (cf.~\cite{goncalvesBiasedRandomkeyGenetic2011}), we thus compose the genotype as $\genotype = [\actionspart_1,\dots,\actionspart_N, \allowbreak \valuespart_{N+1},\dots,\valuespart_{2N}] \allowbreak = [\actionspart, \valuespart]$, with $\actionspart_i, \valuespart_i \in [0,1]$.

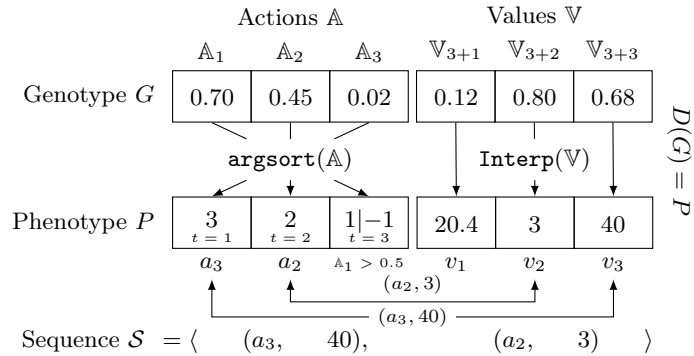
\begin{figure}[!ht]
	\centering
	\begin{tikzpicture}[>=latex]
	\matrix[mymat,anchor=west,nodes={draw}]%,row 2/.style={nodes=draw}]
	at (0,0) 
	(genoAction)
	{
		0.70 & 0.45 & 0.02 \\ 
		%0.12 & 0.80 & 0.69 & 0.73 \\
	};
	\matrix[mymat,right=-1.0ex of genoAction,nodes={draw}]%,row 2/.style={nodes=draw}]
	(genoValues)
	{
		0.12 & 0.80 & 0.68 \\
	};
	
	\matrix[mymat,style={text height=1.5ex},below=5.5ex of genoAction,nodes={draw}]%,row 2/.style={nodes=draw}]
	(phenoAction)
	{
		3 & 2 & 1 \lvert -1\\ 
	};
	\matrix[mymat,below=5.5ex of genoValues,nodes={draw}]%,row 2/.style={nodes=draw}]
	(phenoValues)
	{
		20.4 & 3 & 40 & \\
	};
	
	\node[above=0pt of genoAction-1-1.north]
	(a1) {$\actionspart_1$};
	\node[above=0pt of genoAction-1-2.north]
	(a2) {$\actionspart_2$};
	\node[above=0pt of genoAction-1-3.north]
	(a3) {$\actionspart_3$};
	
	\node[above=0pt of a2]
	(actionsLabel) {Actions $\actionspart$};

	\node[below=-9pt of phenoAction-1-1.south]
	() {\tiny $t=1$};	
	\node[below=-9pt of phenoAction-1-2.south]
	() {\tiny $t=2$};	
	\node[below=-9pt of phenoAction-1-3.south]
	() {\tiny $t=3$};	
	
	\node[below=0pt of phenoAction-1-1.south]
	(aa1) {$\action_3$};
	\node[below=0pt of phenoAction-1-2.south]
	(aa2) {$\action_2$};
	\node[below=0pt of phenoAction-1-3.south]
	(aa3) {\tiny $\actionspart_1 > 0.5$};
	
	\node[above=0pt of genoValues-1-1.north]
	(v1) {$\valuespart_{3+1}$};
	\node[above=0pt of genoValues-1-2.north]
	(v2) {$\valuespart_{3+2}$};
	\node[above=0pt of genoValues-1-3.north]
	(v3) {$\valuespart_{3+3}$};
	
	\node[above=0pt of v2]
	(valuesLabel) {Values $\valuespart$};

	\node[below=0pt of phenoValues-1-1.south]
	(vv1) {$\Value_1$};
	\node[below=0pt of phenoValues-1-2.south]
	(vv2) {$\Value_2$};
	\node[below=0pt of phenoValues-1-3.south]
	(vv3) {$\Value_3$};

	\node[left=0pt of genoAction] 
	(genotypeLabel) {Genotype $\genotype$};
	\node[left=0pt of phenoAction] 
	(phenotypeLabel) {Phenotype $\phenotype$};

    % \path (genoAction-1-2.south) -- node (argsortLab) {\small \texttt{argsort}} (phenoAction-1-2.north);
    % \node[below=0pt of genoAction-1-2.south]
    % (argsortLab) {\small \texttt{argsort}};
    
    % \path (genotypeLabel.south) -- node[right=0pt of genotypeLabel] (argsortLab) {\small $\texttt{argsort}(\actionspart)$} (phenotypeLabel.north);
    \path (genoAction-1-2.south) -- node (argsortLab) {\small $\texttt{argsort}(\actionspart)$} (phenoAction-1-2.north);
    
	%\node%[right=35pt of argsortLab] 
    \path (genoValues-1-2.south) -- node (interpolateLab) {$\texttt{Interp}(\valuespart)$} (phenoValues-1-2.north);

	\node[below=30pt of phenotypeLabel]
	(sequenceLabel) {Sequence $\sequence$};
	
	\node[right=0pt of sequenceLabel]
	(langle) {$= \langle$};
	\node[right=10pt of langle]
	(action3) {$(\action_3,$};
	\node[right=10pt of action3]
	(value3) {$40),$};
	\node[right=40pt of value3]
	(action2) {$(\action_2,$};
	\node[right=10pt of action2]
	(value3) {$3)$};
	\node[right=10pt of value3]
	(rangle) {$\rangle$};

	%	\node[below=10pt of phenoAction-1-3.south]
	%		(actionNan) {$\action_{\emptyset}$};

	\begin{scope}%[shorten <= -2pt]
		%		\draw[->]
		%		(genotypeLabel.west) -- (phenotypeLabel.west);
		
% 		\draw[->]
% 		(genoAction-1-3.south) -- (phenoAction-1-1.north);
% 		\draw[->]
% 		(genoAction-1-2.south) -- (phenoAction-1-2.north);
% 		\draw[->]
% 		(genoAction-1-1.south) -- (phenoAction-1-3.north);
		
		\draw[->]
		(genoValues-1-1.south) -- (phenoValues-1-1.north);
		\draw[->]
		(genoValues-1-2.south) -- (interpolateLab) -- (phenoValues-1-2.north);
		\draw[->]
		(genoValues-1-3.south) -- (phenoValues-1-3.north);

%         % NEW
		\draw[->]
		(genoAction-1-3.south) -- (argsortLab) -- (phenoAction-1-1.north);	
		\draw[->]
		(genoAction-1-2.south)  -- (argsortLab) -- (phenoAction-1-2.north);
		\draw[->]
		(genoAction-1-1.south)  -- (argsortLab) -- (phenoAction-1-3.north);
		
% 		\draw[*->]
% 		(genoValues-1-1.south) -- (phenoValues-1-1.north);
% 		\draw[*->]
% 		(genoValues-1-2.south) -- (phenoValues-1-2.north);
% 		\draw[*->]
% 		(genoValues-1-3.south) -- (phenoValues-1-3.north);
		
% 		\draw[*->,dashed,line width=0.7pt]
% 		(phenoValues-1-3.south) to[out=-90,in=90] (action2Value.north);
% 		\draw[*->,dashed,line width=0.7pt]
% 		(phenoValues-1-2.south) to[out=-90,in=90] (action1Value.north);

	\end{scope}
	
    \node[right=10pt of genoValues-1-3, rotate=-90] (decoderLabel) {$\decoder(\genotype) = \phenotype$};
    
	\begin{scope}
% 		every node/.style={fill=white,circle},]
% 		every edge/.style={draw=black!60,dashed,very thick}]
    \draw[<->]
    (aa1.south) |- ++(0,-5mm) -| (vv3.south) node[pos=0.25, fill=white] {\scriptsize $(\action_3,40)$};
    \draw[<->]
    (aa2.south) |- ++(0,-3mm) -| (vv2.south) node[above, pos=0.25] {\scriptsize $(\action_2,3)$};
    \end{scope}
    
% 	\draw[->]
% 	(phenoAction-1-1.south) -- (action3.north);

% 	\node[style={draw=red, thick, dotted, inner sep=0.5em}, fit=(genoAction-1-1) (genoValues-1-3)] {};
% 	\node[style={draw=blue, thick, dotted, inner sep=0.5em}, fit=(phenoAction-1-1) (phenoValues-1-3)] {};
	%		\node[style={draw=orange, thick, dotted, inner sep=0.5em}, fit=(actionsLabel)(phenoAction-1-1) (phenoAction-1-3)] {};
	%		\node[style={draw=green, thick, dotted, inner sep=0.5em}, fit=(valuesLabel)(phenoValues-1-1) (phenoValues-1-3)] {};
	
\end{tikzpicture}
	\caption{Anatomy and representation of the solution decoding.}
	\label{fig:decoder}
\end{figure}

The first $N$ values, $\actionspart$, in $G$ encode the action subset $\actions$ (cf.~\ref{prob:1}) and their ordering $t \in \{1,\dots,T\}$ in the sequence $\sequence$ (cf.~\ref{prob:3}). Each index position $i \in \{1,\dots,|\actionspart|\}$ corresponds to one of the actions in the action set $\action_i \in \actions$ (i.e. $\actionspart_{i} \in \actionspart$ encodes $\action_i \in \actions$).
The other half, $\valuespart$, encodes the tweaking values $\Values$ (cf.~\ref{prob:2}) of each action, which is also referred to by the index position $i \in \{1,\dots,|\valuespart|\}$ (i.e. $\valuespart_{N+i} \in \valuespart$ encodes $\Value_i \in \Values_h$).
Fig.~\ref{fig:decoder} visualizes this composition of the genotype.

\textbf{Decoder}: \,
Since BRKGA itself is problem-independent, we design a problem-specific \emph{decoder} $\decoder(\genotype) = \phenotype$ to infer the phenotype $\phenotype$ from $\genotype$.
Below we discuss its design, which is inspired by established concepts (cf.~\cite{goncalvesBiasedRandomkeyGenetic2011}).

The subset of actions (cf.~\ref{prob:1}) is decoded by identifying \emph{inactive} actions in the actions part $\actionspart$. As a simple heuristic, we define an action in $\genotype$ as inactive (denoted by \enquote{$-1$}), if its genotype value is greater than $0.5$. This follows from the idea that an action has an equal chance to be active or inactive when chosen randomly.
To get the \emph{active} actions and their \emph{order} (cf.~\ref{prob:3}), we apply the commonly used $\texttt{argsort}: \actionspart \to \actions$ decoding~\cite{goncalvesBiasedRandomkeyGenetic2011,beanGeneticAlgorithmsRandom1994} on $\actionspart$ and identify the inactive actions afterwards, which will always produce a non-repeating order. We find the actual actions by looking at the sorted index (cf. $\phenotype$ in Fig.~\ref{fig:decoder}). Note that an action will be at an earlier position $t$ in $\sequence$ the lower its genotype value is.
Lastly, we decode the values part by applying a (linear) interpolation $\texttt{Interp}: \valuespart \to \Values$ on each of the genotype values in $\valuespart$.
Therefore, only the original value ranges need to be provided (via $\Values$), or in case of categorical values a mapping between the interpolated value and the respective categorical value counterpart. The decoded value at position $i \in \{1,\dots,|\valuespart|\}$ then belongs to $\action_i \in \sequence$.

An example of the full decoding process (from a genotype solution $\genotype$ to the actual solution sequence $\sequence$) is visualized in Fig.~\ref{fig:decoder}.
As we can see, applying \texttt{argsort} on $\actionspart \in \genotype$ realizes an order (3,2,1) for $\actions$ via $\phenotype$. Since $\actionspart_1 > 0.5$, action $\action_1$ is rendered inactive and the remaining, ordered, action set is $\langle \action^1_3, \action^2_2 \rangle$. The associated tweaking values are then decoded by interpolating $\valuespart \in \genotype$ and assigned to their action counterparts, thus creating the sequence $\sequence$.
Note that decoding is necessary for all solutions in each iteration to evaluate the fitness and is repeated until the termination criterion (e.g. maximum number of iterations) is reached.

\section{Experiments}
\label{sec:experiments}
The first goal of our experiments is to evaluate the costs of the generated sequences\footnote{Apart from the cost objective $\seqCosts$, we will not report on the remaining objective space from Eq.~\ref{eq:problem} here, since the results were similar and thus not particularly informative.} in comparison to the state-of-the-art approach~\cite{ramakrishnanSynthesizingActionSequences2020} (Sect.~\ref{sec:exp_competitor_minimal_costs}).
Next, we analyze the diversity of the generated solutions in terms of the action space and the sequence orders (Sect.~\ref{sec:exp_sequence_order}).
Finally, we examine the effect of the action positions in a sequence for switching the class label (Sect.~\ref{sec:exp_action_class_response}).

\textbf{Datasets}: \,
We report on the datasets \adultDataset{}~\cite{duaUCIMachineLearning2017} (for predicting whether income exceeds \$50k/year based on census data) and \germanDataset{}~\cite{duaUCIMachineLearning2017} (for predicting whether a loan application will be accepted).

\textbf{Baselines}: \,
The following three methods were used for the evaluation. The first one acts as a direct competitor and the others are variations of our method.

\begin{itemize}[align=left]
\item[\textbf{\competitor{}}~\cite{ramakrishnanSynthesizingActionSequences2020}:] The competitor\footnote{\url{https://github.com/goutham7r/synth-action-seq}} only optimizes for finding a \emph{single} minimal cost sequence and solves two separate sub-problems independently. First, they generate candidate action subsets according to~\ref{prob:1} with respect to one of their proposed heuristics. Then, they perform an adversarial attack based on~\cite{carliniEvaluatingRobustnessNeural2017} in order to find the tweaking values for each candidate sequence to solve~\ref{prob:2}.
There is no explicit sequence order notion as per~\ref{prob:3} apart from pre- and post-requirements, thus it only optimizes on their provided cost function which is equivalent to the action effort that we introduced in Sect.~\ref{sec:costs}, i.e. $\costs_i = \baseCosts_i$.
To make the comparison to~\cite{ramakrishnanSynthesizingActionSequences2020} fair, we use their exact same action-cost model and provided pre-trained black-box classifiers (which are neural networks here) for all methods.
That means, all action behavior is identical in this comparison (i.e. tweaking effects, constraints, conditions and costs). Hence, we use the costs of their model for the action effort $\baseCosts_i$.
\item[\textbf{\mainEA{}}:] Our method optimizes \emph{all} sub-problems at once and provides \emph{multiple} solutions. Regarding the cost, it considers the consequential discount and thus optimizes $\costs_i = \baseCosts_i \cdot \consequentialCosts_i$.
For $\consequentialCosts_i$, we provided a simple feature relationship graph $\dependencyGraph$ that models beneficial effects in \adultDataset{} such as:
\begin{itemize}
    \item The higher the \texttt{Education} level, the easier it gets to increase \texttt{Capital Gain}, change \texttt{Work-Class} and \texttt{Occupation}.
    \item The lower the \texttt{Work Hours}, the easier it is to increase the \texttt{Education}.
    \item The higher the \texttt{Work Hours}, the easier it gets to increase \texttt{Capital Gain}.
\end{itemize}
We only use \mainEA{} for \adultDataset{} as it was not practical to create a meaningful graph based on the predefined actions from~\cite{ramakrishnanSynthesizingActionSequences2020} for \germanDataset{}.
Since the $\consequentialCosts_i$ part primarily affects the order of actions, we would generally expect \mainEA{} to behave similarly to \alternativeEA{} in terms of $\baseCosts_i$, though.
% The meaningfulness of the sequence should improve, however.
\item[\textbf{\alternativeEA{}}:] This is a variation of our proposed \mainEA{}, leaving out the consequential discount and thus only optimizes on the action efforts from~\cite{ramakrishnanSynthesizingActionSequences2020}, i.e. $\costs_i = \baseCosts_i$.
When referring to findings that apply to both \mainEA{} and \alternativeEA{}, we use \enquote{\oursBoth{}}.
\end{itemize}

\textbf{Implementations}: \,
We implemented our method in Python\footnote{\url{https://github.com/ppnaumann/CSCF}} using the \emph{pymoo}~\cite{blankPymooMultiobjectiveOptimization2020} implementation of BRKGA.
The parameters for BRKGA are mostly based on recommendations from~\cite{goncalvesBiasedRandomkeyGenetic2011}: the population size was set to $500$, the mutant and offspring fractions to $0.2$ and $0.8$, respectively, and the crossover bias to $0.7$. 
As the termination criterion for \oursBoth{}, we fixed the number of iterations to $150$.
From each dataset we chose $100$ random instances that are currently classified by the black-box as the undesired class (i.e. $\texttt{Salary} < \$50\text{k}$ and $\texttt{Credit denied}$) with the intention of generating a sequential counterfactual to flip their class label. Each instance represents an experiment.
We ran all methods on the same $100$ experiments and fixed the maximum sequence length of \competitor{} to $T=2$ because of long runtimes for larger values which made the experiments of those unfeasible on our hardware\footnote{All experiments (competitor and our method) were executed on the free tier of \emph{Google Colab} (\url{https://colab.research.google.com/}).}.
The long runtimes of \competitor{} were already mentioned in their paper: \enquote{Time/iteration is $\sim$15s across instances}~\cite{ramakrishnanSynthesizingActionSequences2020}, which confirms our observations, since the algorithm may take up to a few 100 iterations according to~\cite{ramakrishnanSynthesizingActionSequences2020} (running \competitor{} for $T \leq 2$ took the same time as \oursBoth{} needed for all sequence lengths simultaneously). Because of this, we used the \enquote{\emph{Vanilla}} heuristic for \competitor{} as it was found to perform the best for shorter sequences based on~\cite{ramakrishnanSynthesizingActionSequences2020}.
Lastly, we had to filter out some experiments in the post-processing since \competitor{} produced constraint violating solutions or did not find a feasible one. Consequently, the number of experiments for \germanDataset{} was post-hoc decreased to 85, but for \adultDataset{} it did not change.

\subsection{Sequence Costs of Sequential Counterfactuals}
\label{sec:exp_competitor_minimal_costs}
\begin{figure}[!ht]
    \centering
    \includegraphics[width=\textwidth]{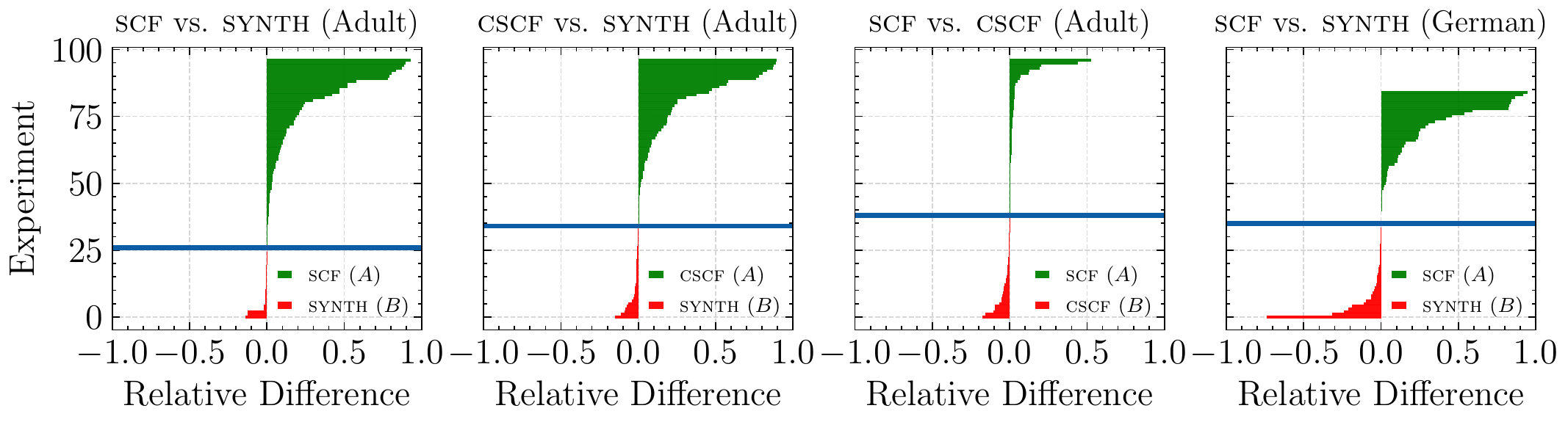}
    \caption{Relative minimal sequence cost ($\seqCosts$) differences between the three methods for both datasets and solutions with $T \leq 2$. It is computed as: $(B-A)/\max{\{A,B\}}$.}
    \label{fig:cost_comparison}
\end{figure}

We show the \emph{undiscounted} (i.e. only using the action effort share $\baseCosts_i$) sequence costs of the solutions ($\seqCosts$ objective) in Fig.~\ref{fig:cost_comparison} for \adultDataset{} and \germanDataset{}.
In the $x$-axis we see the individual, pair-wise relative differences between the computed minimal cost sequences for two methods and each of the valid initial inputs/experiments (which are represented by the $y$-axis).
The green color indicates that the method mentioned first in the title~($A$) performed better, whereas red indicates the other one~($B$) did.
The blue line shows the point from which one method consistently outperforms the other.

Since our method finds \emph{multiple} optimal sequences (on median~4 for \germanDataset{} and~7 for \adultDataset{}) of different lengths per experiment, and \competitor{} only finds a single one, we chose the least cost sequence in \oursBoth{} per set that satisfies $T \leq 2$ in order to make the comparison fair. Note, that there could be a longer sequence with less overall cost that was found by \oursBoth{} due to using more, but cheaper actions.
Although \mainEA{} has optimized on the discounted cost, we only use the effort share, $\baseCosts_i$, of it in this analysis to guarantee comparability.

As we can see in Fig.~\ref{fig:cost_comparison}, \oursBoth{} usually performed better in \adultDataset{}. In \germanDataset{} it seems to be fairly even, but with a slight tendency towards \alternativeEA{}. 
The overall larger relative differences in favor of \oursBoth{} (green), with respect to \competitor{}, appear to be the result of \competitor{} selecting a different, but more expensive set of actions. By looking through the history trace, we identified that the same set of actions that \oursBoth{} found to be optimal was evaluated by \competitor{}, although with different tweaking values. These values, however, seemed to produce a constraint-breaking solution that was either rendered invalid by \competitor{}, or had high costs, since constraints are enforced as a penalty in \competitor{}.
The cases where \competitor{} outperforms \oursBoth{} (red) show small cost differences only. Notably, the differences between \mainEA{} and \alternativeEA{} are also minor, even though \mainEA{} optimizes on the discounted costs.
Thus, this suggests that the augmentation by $\consequentialCosts_i$ does not significantly interfere with the goal of keeping $\totalCosts_\sequence$ minimal.
In general, we conclude that \oursBoth{} is capable to find equivalent or better solutions in comparison to \competitor{} in terms of costs.

\subsection{Diversity of Sequential Counterfactuals}
\label{sec:exp_sequence_order}
\begin{figure}[!ht]
	\subfloat[\alternativeEA{} (A)]{\includegraphics[width = 0.5\textwidth]{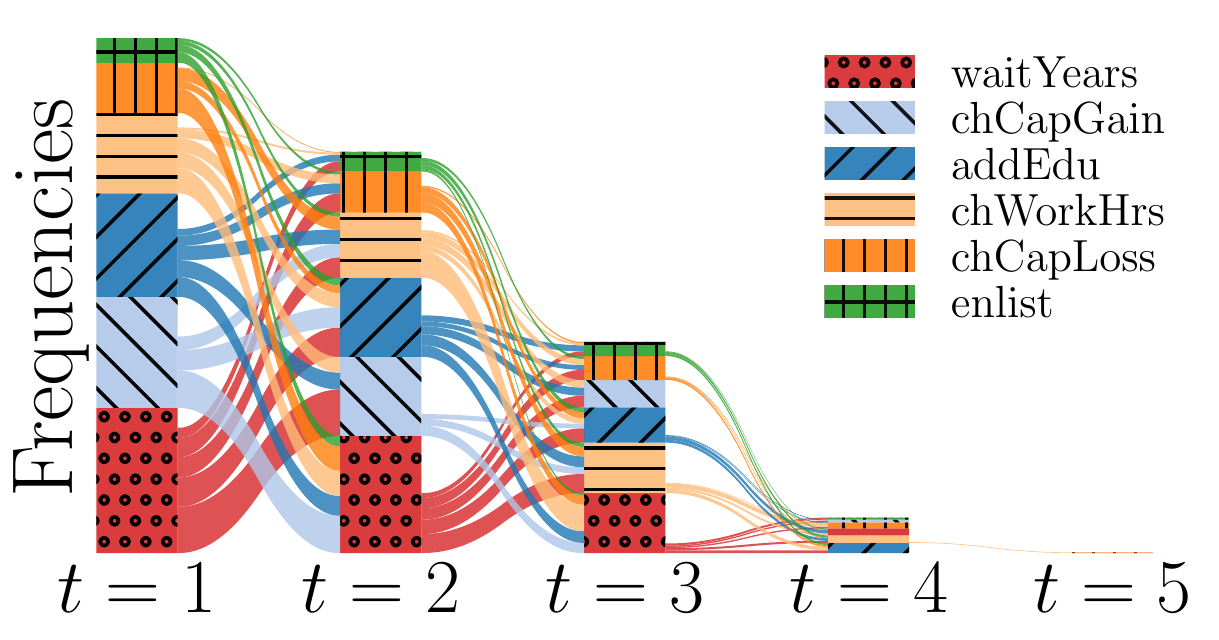}} 
	\subfloat[\mainEA{} (A)]{\includegraphics[width = 0.5\textwidth]{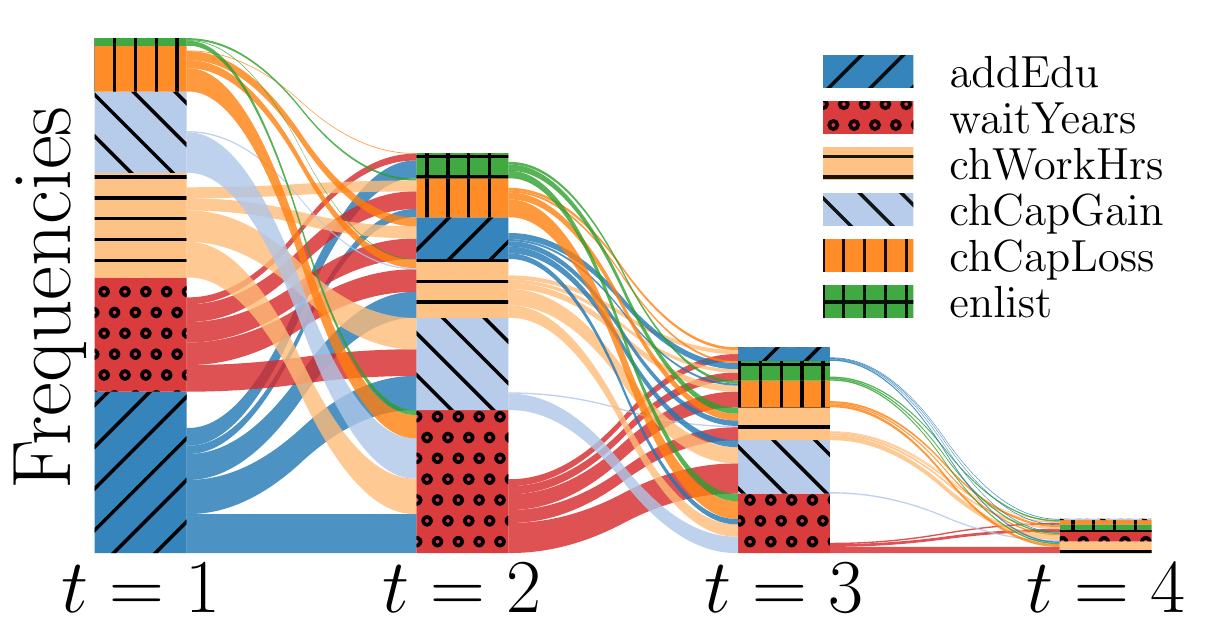}}
	
	\subfloat[\alternativeEA{} (A) ($T \leq 2$)]{\includegraphics[width = 0.33\textwidth]{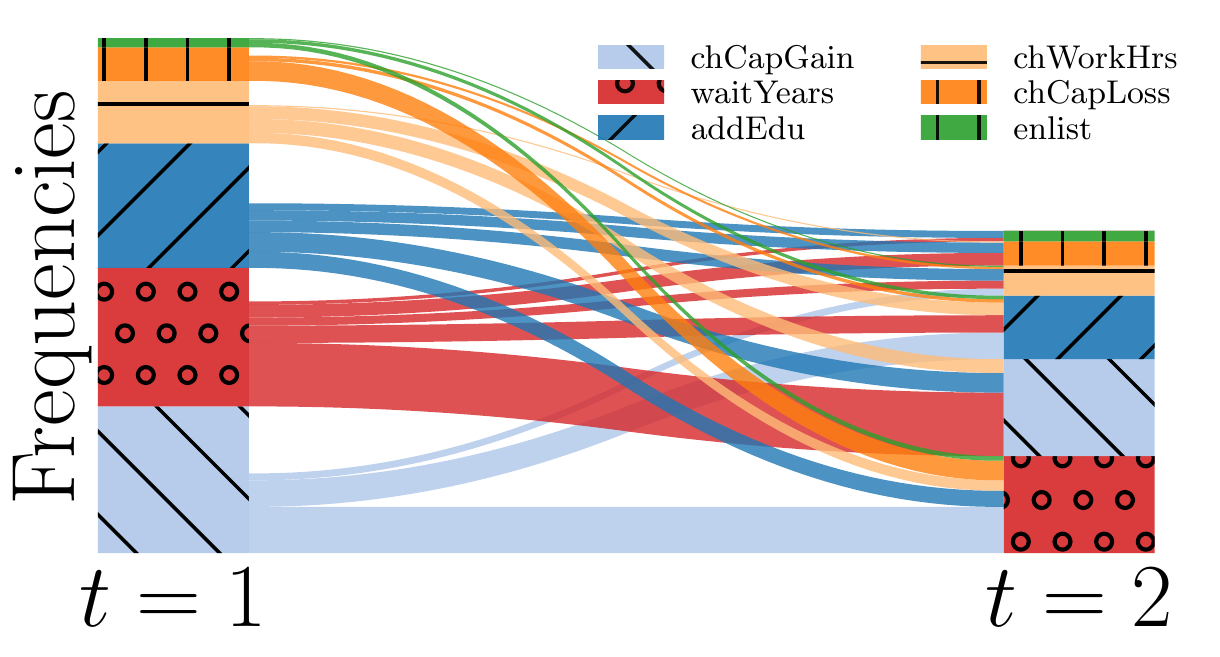}}
	\subfloat[\mainEA{} (A) ($T \leq 2$)]{\includegraphics[width = 0.33\textwidth]{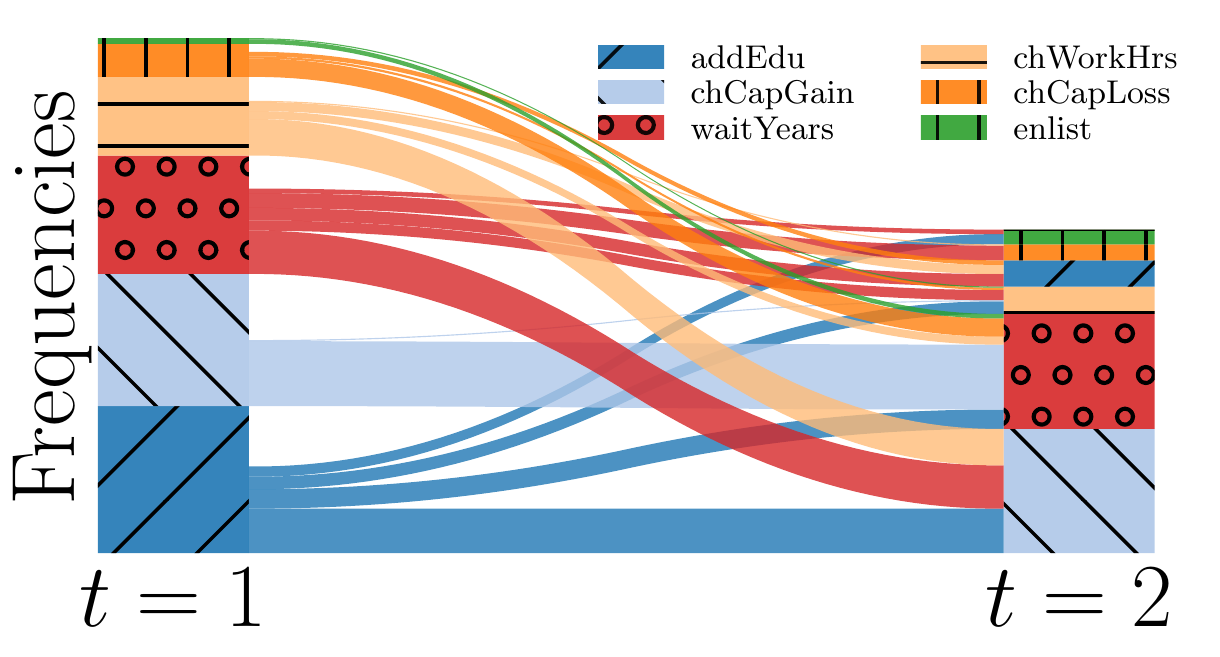}}
	\subfloat[\competitor{} (A) ($T \leq 2$)]{\includegraphics[width = 0.33\textwidth]{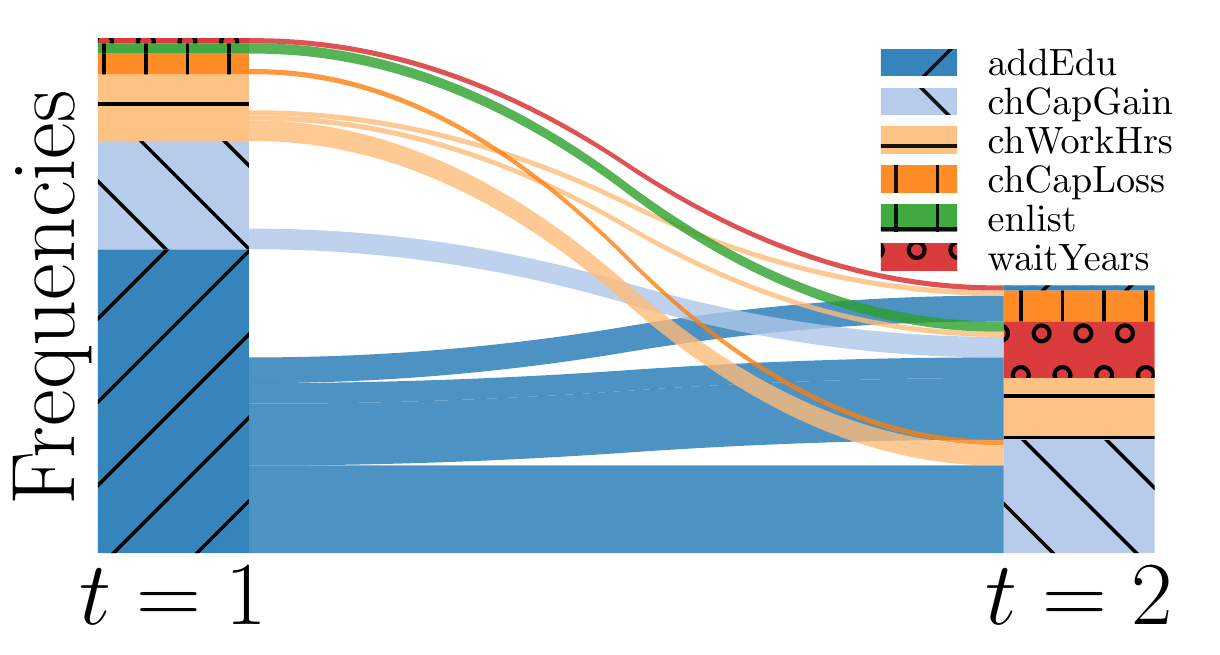}}
	
	\subfloat[\alternativeEA{} (B)]{\includegraphics[width = 0.33\textwidth]{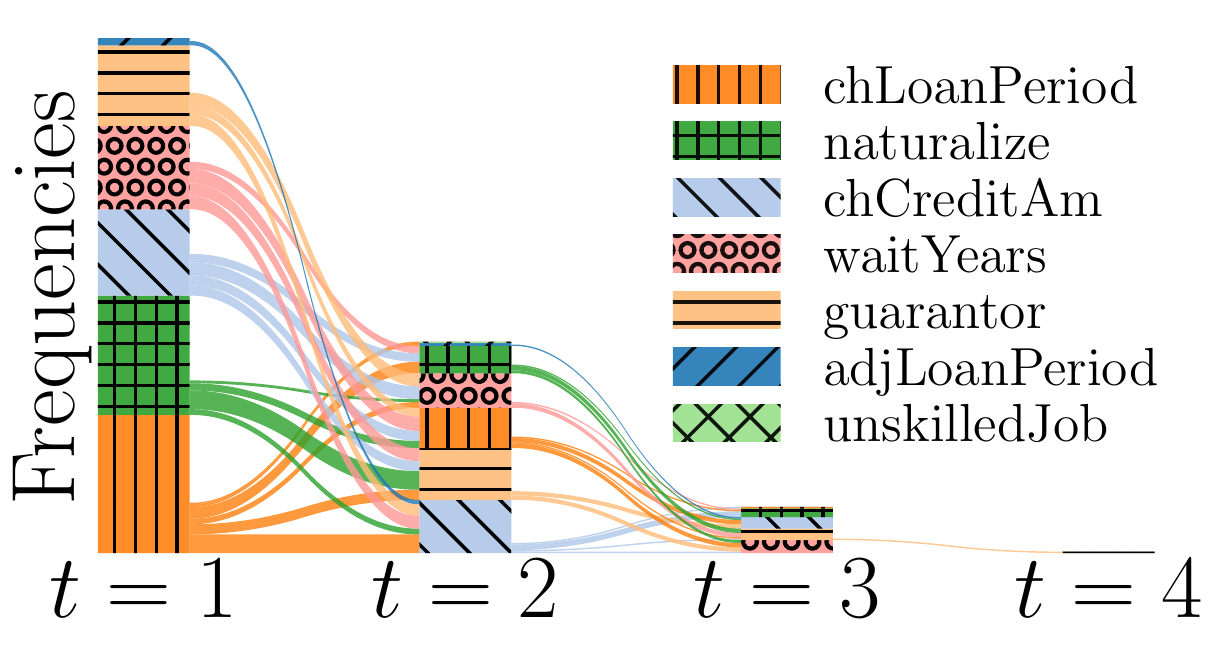}} 
	\subfloat[\alternativeEA{} (B) ($T \leq 2$)]{\includegraphics[width = 0.33\textwidth]{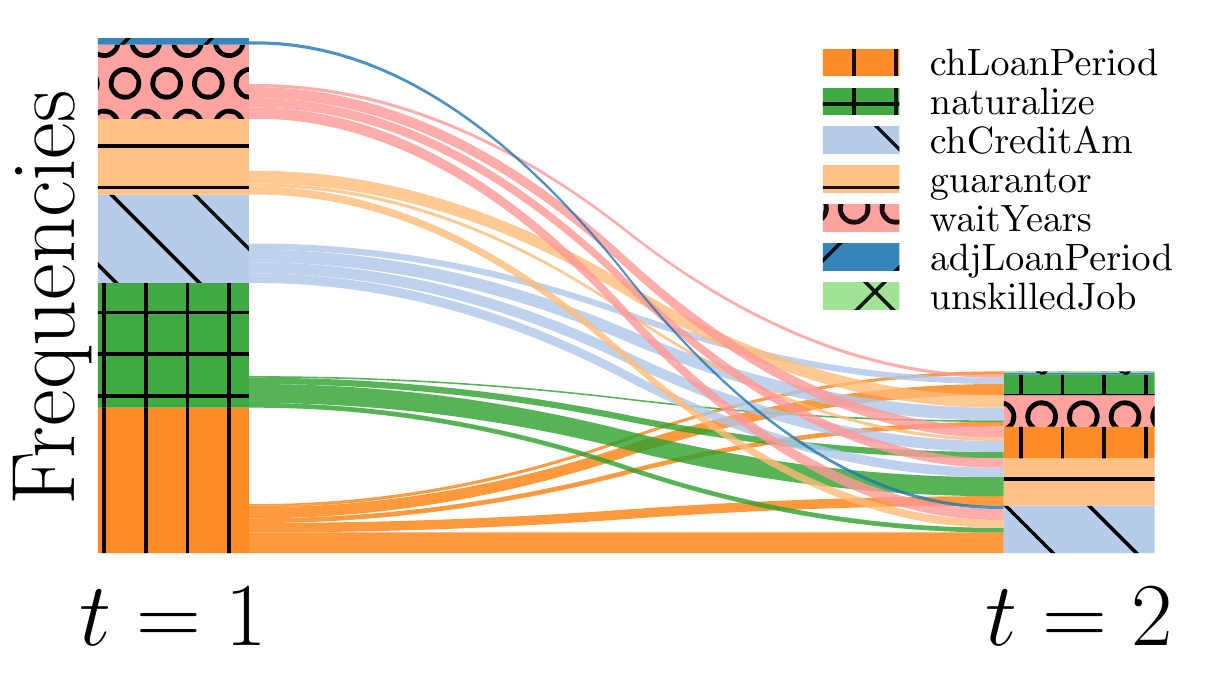}}
	\subfloat[\competitor{} (B) ($T \leq 2$)]{\includegraphics[width = 0.33\textwidth]{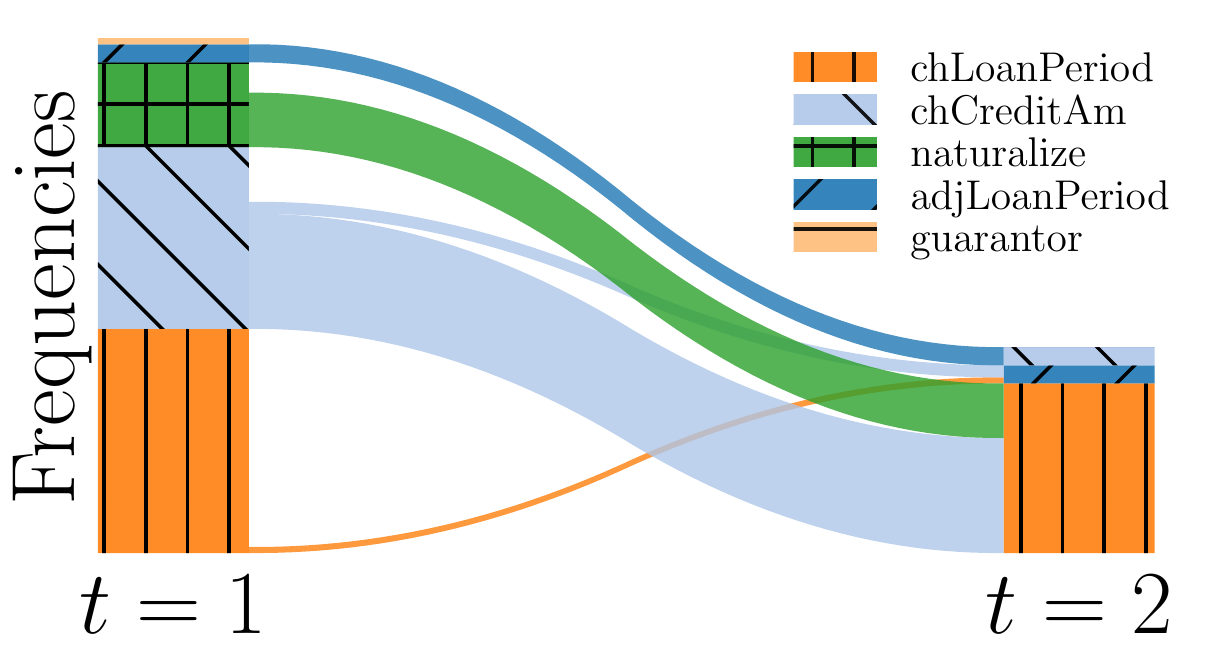}} 
    
    \caption{Sankey plots showing the sequence orders and flow between subsequent actions for differently aggregated solution (sub-)sets. The first two rows correspond to \adultDataset{} (A) and the third to \germanDataset{} (B).}
	\label{fig:sequence_flows}
\end{figure}

In Fig.~\ref{fig:sequence_flows} we illustrated the prevalence of actions at different positions $t$ in a sequence (indicated by the height of each action in the pillars) along with the frequency of how often one action followed on from another (indicated by the widths of the flows). The whitespace of an action in the pillar shows that a sequence stopped there (i.e. had no subsequent actions).
For this purpose, we aggregated over \emph{all} optimal solution sequences (i.e. the whole final Pareto-fronts) from each experiment per method and dataset, respectively. Furthermore, we additionally show \oursBoth{} after filtering out all solutions with $T > 2$~(c,d,g in Fig.~\ref{fig:sequence_flows}) for better comparability with \competitor{}.
The plots~(a,b,c,d,e) in Fig.~\ref{fig:sequence_flows} belong to the \adultDataset{}~(A) and (f,g,h) to the \germanDataset{}~(B) dataset. 

As we can see in Fig.~\ref{fig:sequence_flows}~(a,b,c,d,f,g), \oursBoth{} makes use of the whole available action space $\actions$, and more evenly utilizes the actions in each step than \competitor{}~(e,h). For \germanDataset{}, we notice that \competitor{} only used five different actions, whereas \alternativeEA{} used all available (f,g). This observation is in alignment with our diversity goal and can be attributed to the tweaking frequency objectives ($\objective_{2+h}$ from Eq.~\ref{eq:problem}) which force the algorithm to seek alternative sequences that propose different changes while still providing minimal costs. Since \competitor{} only finds the single least cost solution, the same actions were chosen as they appear to be less expensive than their alternatives.
Regarding the lengths of the found sequences, we see that the minimal cost ones were usually found up to length three according to (a,b,f). After that, only few sequences still provide some sort of minimal cost. Because of this, we can say that \oursBoth{} does implicitly favor shorter sequences if they are in alignment with the costs. This behavior also follows from the tweaking frequency objectives, which minimize the number of times a feature was changed and thus the number of actions used (as these are directly related to another).

Looking at the particular differences between \mainEA{}~(b,d) and \alternativeEA{}~(a,c), we can identify the distinct characteristics of discounting each action effort by $\consequentialCosts_i$ through $\dependencyGraph$. The relationship \enquote{the higher the \texttt{Education}, the easier it gets to attain \texttt{Capital Gain}} is reflected in (b,d) as \texttt{addEdu} is the most frequent action at $t=1$. Moreover, there is no single sequence where \texttt{addEdu} would appear \emph{after} \texttt{chCapGain}, indicating that the beneficial consequence was always used by \mainEA{}. In comparison, \alternativeEA{} in (a,c) has a more equal spread as it has no knowledge of the relationships. Lastly, the same peculiarity can be observed for \texttt{chWorkHrs}, which was more often favored to be placed before \texttt{chCapGain} in \mainEA{} than \alternativeEA{} because of the beneficial relation in $\dependencyGraph$.
Even though it appears that the \texttt{addEdu} effect is also visible in \competitor{} according to (e), this is an artifact since there is no explicit mechanism that would enforce it.
The most likely reason for this is the order in which the actions were processed in the \emph{Vanilla} heuristic.

Finally, looking at the $T \leq 2$ plots~(c,d,e,g,h) specifically, we can see that some actions show a preferred co-occurrence. E.g., \texttt{chCapGain} and \texttt{waitYears} seem to appear more often subsequently than others (c,d). This is not visible in \competitor{}~(e), which on the other hand shows a distinct co-occurrence of \texttt{chCreditAm} and \texttt{chLoanPeriod}.
The reason for this can be traced back to the cost model, which values these combinations as least expensive for sequence lengths of $T \leq 2$ (i.e. if we only use two actions). In case of \oursBoth{} this effect is weaker though, as it seeks for alternatives by design (cf. diversity principle from Sect.~\ref{sec:moo_problem}).

\subsection{Effect of the Action Positions on Achieving the Counterfactual}
\label{sec:exp_action_class_response}
\begin{figure}[!ht]
    \centering
    \includegraphics[width=\textwidth]{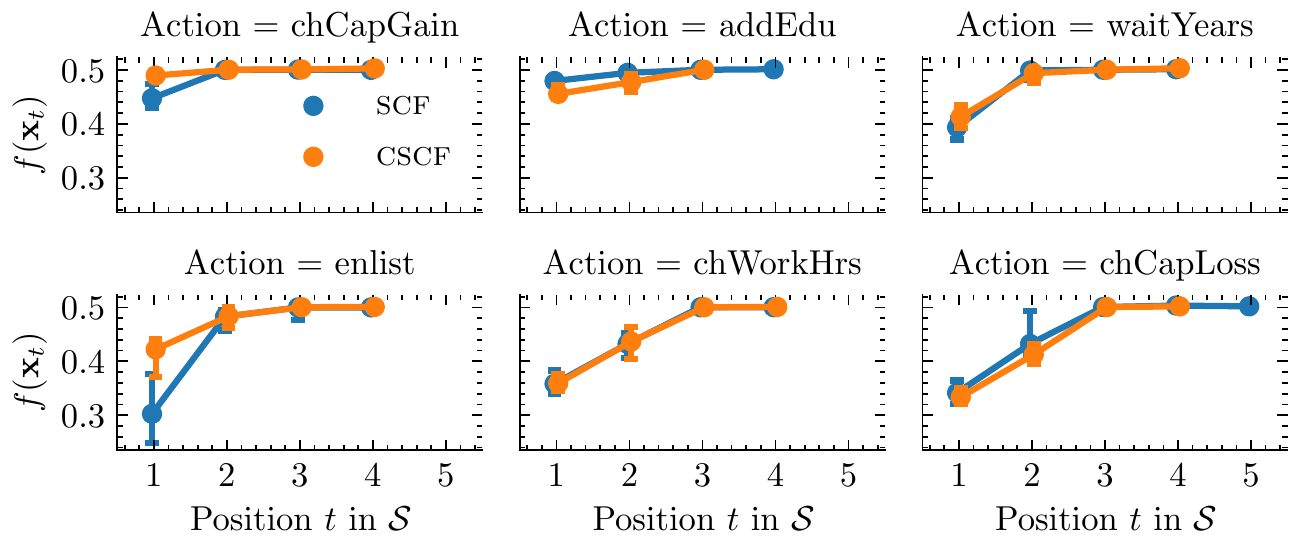}
    \caption{Median effect with 95\% confidence interval of each action at position $t$ in a sequence on the target class (\targetclass{}) probability for \adultDataset{}. $\blackbox(\state_t) > 0.5$ indicates the class label switched at position $t$.}
    \label{fig:adult_probabilities}
\end{figure}

Lastly, looking at Fig.~\ref{fig:adult_probabilities} we can see how each action affects the target class probability of \targetclass{} in the \adultDataset{} with respect to their positional occurrence in a sequence. We again aggregated over \emph{all} computed solutions here.
The $x$-axis denotes the position $t$ in the sequence and the $y$-axis shows the median probability of the target class based on the black-box and the bootstrapped $95\%$ confidence interval (i.e. $2.5$ \& $97.5$ percentiles). 
% Here, $\blackbox(\state_t) > 0.5$ means the class label switched.

As we can see, there are some actions that are almost able to switch the class on their own when used (\texttt{chCapGain}, \texttt{addEdu}), whereas the remaining ones only do it later at position two (\texttt{waitYears}, \texttt{enlist}) or three (\texttt{chWorkHrs}, \texttt{chCapLoss}). 
Based on this, it suggests that some actions are only of supportive nature, whereas others can be seen as the main driver behind class label changes.
Additionally, the actions that affect a feature, which was attributed a consequential effect through $\dependencyGraph$ (\texttt{chCapGain}, \texttt{addEdu}, \texttt{enlist}), are the only ones that show a significant difference here. 
Furthermore, the effect of $\dependencyGraph$ in \mainEA{} is visible. When $\dependencyGraph$ was used, \texttt{chCapGain} changed the class often on position one already, whereas in \alternativeEA{} it was usually not quite possible. Based on this, we can infer that \texttt{chCapGain} at position one in \mainEA{} was typically used when it was able to change the class on its own.
Moreover, it suggests that the \texttt{Education} level was already sufficiently high in the input instance, so that \texttt{chCapGain} was able to increase more while benefiting from the discount enough that the costs were kept low.
The same might be the reason for \texttt{enlist} (i.e. joining the army), which was more able to change the class in \mainEA{}. Since this action affects the \texttt{Occupation} feature, the beneficial edge of \texttt{Education} might have discounted the cost here again so much that it became a minimal cost sequence, whereas in \alternativeEA{} it might have been more expensive. The slightly lower probability of \texttt{addEdu} in \mainEA{} may further suggest that \texttt{Education} was more commonly used as a supporting action in order to discount future actions (hence its disappearance after $t=3$).

In summary, we saw that \oursBoth{} is able to find more diverse sequences while asserting the least cost objective with comparable or better performance to \competitor{} and being more efficient. Furthermore, we demonstrated that the usage of $\dependencyGraph$ in \mainEA{} provides the desired advantage of more meaningful sequence orders, according to the feature relationships, while maintaining minimal costs.

\section{Conclusion and Future Work}
\label{sec:conclusion}
We proposed a new method \mainEA{} for sequential counterfactual generation that is  model-agnostic and capable of finding multiple optimal solution sequences of varying sequence lengths.
Our variants, \mainEA{} and \alternativeEA{}, yield better or equivalent results in comparison to \competitor{}~\cite{ramakrishnanSynthesizingActionSequences2020} while being more efficient.
Moreover, our extended consequence-aware cost model in \mainEA{}, that considers feature relationships, provides more meaningful sequence orders compared to \alternativeEA{} and \competitor{}~\cite{ramakrishnanSynthesizingActionSequences2020}.
In future work, we aim to incorporate causal models to estimate consequential effects in the feature and cost space. Additionally, we want to investigate alternative measures and objectives for evaluating sequence orders and develop a final selection guide for the end user for choosing a sequence from the solution set.
\bibliographystyle{splncs04}
% \bibliography{bibliography}

%
%
%
\end{document}